%% file: main.tex
\documentclass[10pt,twocolumn,letterpaper]{article}

\usepackage[pagenumbers]{iccv} 

\input{preamble}

\definecolor{iccvblue}{rgb}{0.21,0.49,0.74}
\usepackage[pagebackref,breaklinks,colorlinks,allcolors=iccvblue]{hyperref}

\def\paperID{13610} 
\def\confName{ICCV}
\def\confYear{2025}

\title{MEMFOF: High-Resolution Training for \\Memory-Efficient Multi-Frame Optical Flow Estimation}

\author{
Vladislav Bargatin$^{1}$
\quad
Egor Chistov$^{1}$
\quad
Alexander Yakovenko$^{1,2}$
\quad
Dmitriy Vatolin$^{1,2}$
\vspace{0.3em} \\
{\normalsize $^1$\textit{Lomonosov Moscow State University, Moscow, Russia}} \\
{\normalsize $^2$\textit{MSU Institute for Artificial Intelligence, Moscow, Russia}} \\
{\small \texttt{\{vladislav.bargatin, egor.chistov, alexander.yakovenko, dmitriy\}@graphics.cs.msu.ru}}
}

\begin{document}
\maketitle
\input{0_abstract}
\input{1_intro}
\input{2_relatedworks}

\input{3_method}

\input{4_experiments}
\input{5_conclusion}
\input{6_acknowledgments}

\clearpage

{
    \small
    \bibliographystyle{ieeenat_fullname}
    \bibliography{main}
}

\input{X_suppl}

\end{document}

%% file: preamble.tex
\usepackage{multicol}
\usepackage{multirow}
\usepackage{flushend}
\usepackage{array}
\usepackage{tabularray}

\definecolor{lightgray}{rgb}{0.85,0.85,0.85}

\usepackage{booktabs,arydshln}

\makeatletter
\def\adl@drawiv#1#2#3{%
        \hskip.5\tabcolsep
        \xleaders#3{#2.5\@tempdimb #1{1}#2.5\@tempdimb}%
                #2\z@ plus1fil minus1fil\relax
        \hskip.5\tabcolsep}
\newcommand{\cdashlinelr}[1]{%
  \noalign{\vskip\aboverulesep
           \global\let\@dashdrawstore\adl@draw
           \global\let\adl@draw\adl@drawiv}
  \cdashline{#1}
  \noalign{\global\let\adl@draw\@dashdrawstore
           \vskip\belowrulesep}}
\makeatother

%% file: 0_abstract.tex
\begin{abstract}

Recent advances in optical flow estimation have prioritized accuracy at the cost of growing GPU memory consumption, particularly for high-resolution (FullHD) inputs. We introduce MEMFOF, a memory-efficient multi-frame optical flow method that identifies a favorable trade-off between multi-frame estimation and GPU memory usage. Notably, MEMFOF requires only 2.09 GB of GPU memory at runtime for 1080p inputs, and 28.5 GB during training, which uniquely positions our method to be trained at native 1080p without the need for cropping or downsampling.

We systematically revisit design choices from RAFT-like architectures, integrating reduced correlation volumes and high-resolution training protocols alongside multi-frame estimation, to achieve state-of-the-art performance across multiple benchmarks while substantially reducing memory overhead. Our method outperforms more resource-intensive alternatives in both accuracy and runtime efficiency, validating its robustness for flow estimation at high resolutions. At the time of submission, our method ranks first on the Spring benchmark with a 1-pixel (1px) outlier rate of 3.289, leads Sintel (clean) with an endpoint error (EPE) of 0.963, and achieves the best Fl-all error on KITTI-2015 at 2.94\%.
The code is available at: \url{https://github.com/msu-video-group/memfof}.
\end{abstract}

%% file: 1_intro.tex
\section{Introduction}
\begin{figure}[t]
    \includegraphics[width=\columnwidth]{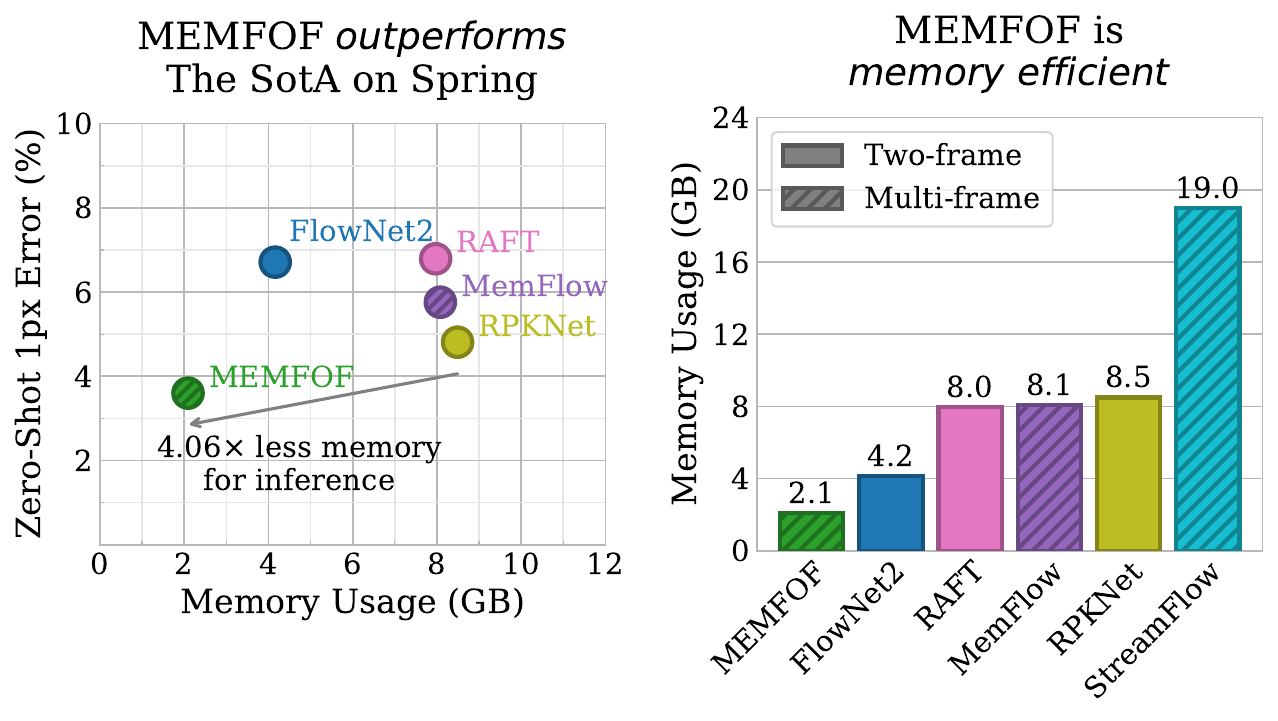}
    \caption{Comparison with state-of-the-art optical flow methods. \textbf{Left}: Quality-memory trade-off on the Spring~\cite{mehl2023spring} benchmark. MEMFOF demonstrates superior memory efficiency and the lowest zero-shot error among all methods. \textbf{Right}: GPU memory consumption for 1080p resolution inputs. MEMFOF outperforms state-of-the-art methods such as RPKNet~\cite{morimitsu2024recurrent} and StreamFlow~\cite{sun2025streamflow}. For more details please see Tab.~\ref{tab:spring-comparison}.
    StreamFlow~\cite{sun2025streamflow} is omitted from the left plot due to space constraints.}
    \label{fig:sota}
\end{figure}
Optical flow (the dense per-pixel motion between frames) estimation is a fundamental task in low-level vision with wide applications from video action recognition and object detection~\cite{piergiovanni2019representation, sun2018optical, zhao2020improved} to video restoration and synthesis \cite{huang2022real, liu2020video, xu2019quadratic, chan2021basicvsr}. Traditional variational methods formulated flow as an optimization problem~\cite{horn1981determining, lucas1981iterative}, but often struggled with large motions and real-time performance. The deep learning era sparked a leap in both accuracy and processing speed: FlowNet~\cite{dosovitskiy2015flownet} pioneered this shift, with PWC-Net~\cite{sun2018pwc} introducing cost volume warping for efficiency. RAFT~\cite{teed2020raft} later established state-of-the-art accuracy through iterative GRU-based refinement of a 4D all-pairs correlation volume. However, RAFT's quadratic memory scaling with image size creates prohibitive costs for high-resolution inference (e.g., 8 GB at FullHD and 25+ GB at WQHD), forcing input downsampling that degrades motion boundary details.

Recent advances address these limitations through two complementary strategies: (1) enhancing correlation efficiency and (2) exploiting multi-frame temporal cues. Memory-efficient architectures reduce correlation costs via sparse candidate matching (SCV)~\cite{jiang2021learning_SCV}, 1D motion decomposition (Flow1D)~\cite{xu2021high}, or hybrid volumes (HCV)~\cite{zhao2024hybrid}. Transformer-based methods like GMFlow~\cite{xu2022gmflow} and FlowFormer++~\cite{shi2023flowformer++} enable global feature matching with fewer iterations. Multi-frame approaches such as VideoFlow~\cite{shi2023videoflow} and MemFlow~\cite{dong2024memflow} leverage temporal consistency to resolve occlusions, while StreamFlow~\cite{sun2025streamflow} optimizes spatiotemporal processing efficiency.

Despite this progress, high-resolution benchmarks like Spring~\cite{mehl2023spring} remain a challenge. Some methods either downsample inputs ~\cite{wang2025sea} or employ tiling strategies~\cite{weinzaepfel2023croco} to reduce memory consumption, trading it for less accuracy or longer inference. And methods operating at native resolutions tend to use large amounts of memory, prohibiting their use on consumer grade hardware, see Fig. \ref{fig:sota} for details.

In this work, we propose MEMFOF, the first multi-frame optical flow method designed for memory efficiency at FullHD. MEMFOF can be trained and run on full 1080p frames without any downsampling or tiling, using only a few GB of memory at inference – all while achieving state-of-the-art accuracy. 
To achieve this, we extend SEA-RAFT, a two-frame optical flow architecture to incorporate a three-frame strategy. Crucially, we adjust the RAFT-style architecture to drastically cut memory usage (about 4$\times$ lower, down to just 2.09 GB) while enabling multi-frame input, allowing our model to run at 1080p on common GPUs. Which in turn allows for training at native 1080p using under 32GB of memory. 

For better handling of large motions found at high resolutions, we devise a training regime that overcomes the mismatch between standard optical flow datasets (often limited in image size and motion magnitude) and the FullHD domain by upscaling existing datasets and training at higher resolutions, see Figure~\ref{exp:abl:mv_hist}. An ablation study shows that this upsampling is critical to avoid \textit{underfitting} on large-motion regions, leading to consistent performance gains on real high-resolution benchmarks. Notably, our method ranks first at zero-shot evaluation on the Spring benchmark, surpassing all other published work (both zero-shot and fine-tuned in Spring).
To the best of our knowledge, we are the first to address the issues of memory consumption of multi-frame methods at high-resolutions in a principled manner.

\noindent In summary, our key contributions are:
\begin{itemize}
    \item \textbf{Memory-Efficient Multi-Frame Design.} We propose a refined multi-frame RAFT-style architecture that processes FullHD inputs natively, reducing GPU memory needs by up to 3.9$\times$ compared to RAFT / SEA-RAFT, requiring only $\sim$2 GB of GPU memory at 1080p inference, well within the capacity of consumer-grade GPUs.
    \item \textbf{High-Resolution Training Strategy.} A novel FullHD-centric data augmentation and multi-stage learning approach to accurately capture large motions, preventing the underfitting that commonly arises when transferring from low-resolution to high-resolution tasks.
    \item \textbf{State-of-the-Art Results on Multiple Benchmarks.} MEMFOF achieves top accuracy on multiple benchmarks with substantially lower memory overhead. It leads on  Spring~\cite{mehl2023spring}, KITTI-2015~\cite{menze2015object}, and Sintel~\cite{butler2012naturalistic}.
\end{itemize}

%% file: 2_relatedworks.tex
\begin{figure*}[t]
    \begin{minipage}[c]{0.75\linewidth}
    \includegraphics[width=\linewidth]{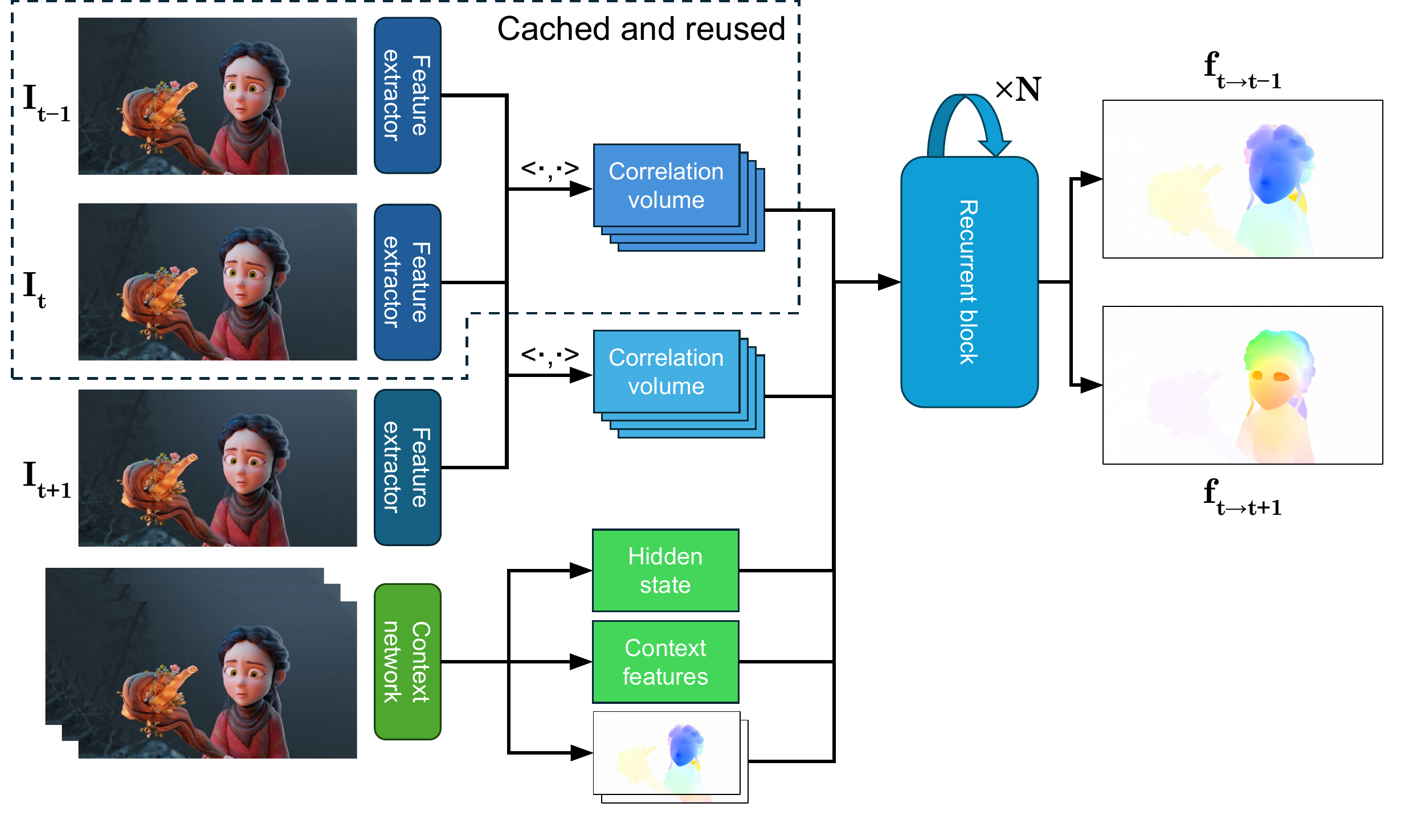}
    \end{minipage}
    \begin{minipage}[c]{0.24\linewidth}
    \includegraphics[width=\linewidth]{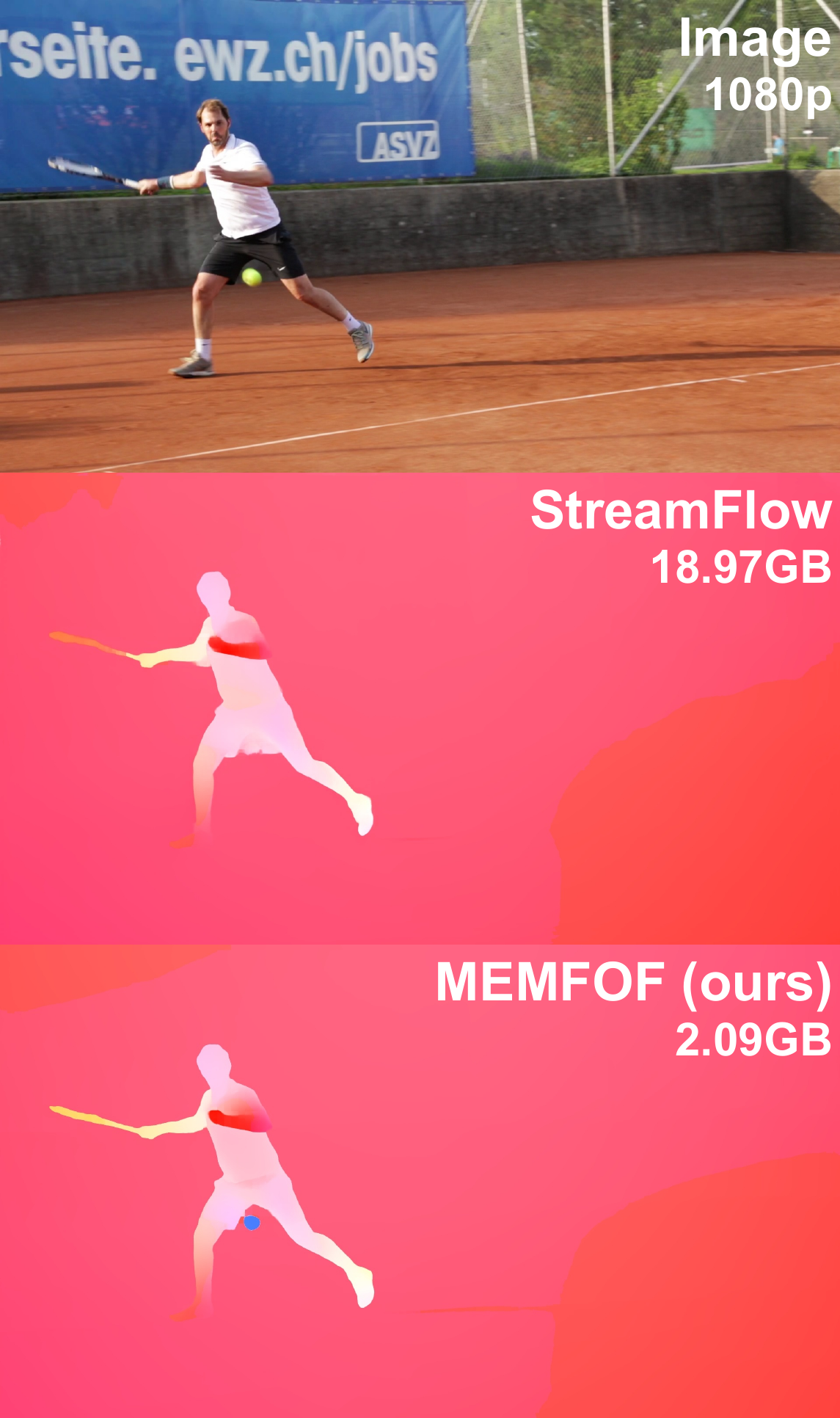}
    \end{minipage}
    \caption{Overview of our method and FullHD inference results. \textbf{Left}: Outline of MEMFOF: when operating on videos we cache and reuse results of the feature extraction stage and correlation volume calculation. For each new frame we extract features and run the context network on the frame triplet, which returns the initial flow estimates, context features and hidden (recurrent) state. The flows are recurrently updated for N iterations and finally upsampled to get the final predictions. \textbf{Right}: Comparison of our method (MEMFOF) with StreamFlow~\cite{sun2025streamflow} on FullHD images from the DAVIS dataset~\cite{perazzi2016benchmark}. Our method correctly captures the tennis ball's movement while requiring much less memory.}
    \label{fig:method}
\end{figure*}

\section{Related Works}

Optical flow estimation is a fundamental problem in computer vision, with applications ranging from motion analysis to video compression. Over the years, various approaches have been proposed to address the challenges of accuracy and efficiency. In this section, we review the existing literature, categorizing it into three main areas: two-frame optical flow, multi-frame optical flow, and memory-efficient optical flow.

\noindent\textbf{Two-frame optical flow.} Classical approaches~\cite{horn1981determining, lucas1981iterative, farneback2003two} optimize an energy function combining similarity and smoothness terms. With the advent of deep learning, methods like FlowNet~\cite{dosovitskiy2015flownet} revolutionized the field by leveraging convolutional neural networks to directly predict optical flow from image pairs. PWC-Net~\cite{sun2018pwc} then introduced a pyramid, warping, and cost volume mechanism.

More recently, RAFT~\cite{teed2020raft} has introduced a new paradigm, employing an iterative refinement process and an all-pair correlation volume. Building on RAFT’s success, several variants have been proposed to improve its efficiency and accuracy. One strategy is to introduce global receptive fields via transformers or attention. GMFlow~\cite{xu2022gmflow} treats optical flow as a global feature matching problem, while FlowFormer~\cite{huang2022flowformer} integrates a transformer into the cost volume processing. Beyond transformers, GMA~\cite{jiang2021learning_GMA} introduces global motion attention to focus the iterative updates on important regions. On the other hand, SEA-RAFT~\cite{wang2025sea} aims to enhance RAFT by three simple tricks: using a mixture of Laplace loss, directly regressing initial flow, and pre-training on a rigid-flow dataset.

Unfortunately, all RAFT-like methods require substantial memory resources on high-resolution inputs. As a result, they are often applied to downscaled frames or with a tiling-based approach, compromising the quality of the estimated flow by losing fine details or global motion context, respectively.

\noindent\textbf{Multi-frame optical flow.} 
While two-frame methods have advanced significantly, they inherently ignore the rich temporal information available in video streams. Early multi-frame attempts simply extended two-frame methods with flow propagation, for example, by fusing the backward warped past flow with current flow through a fusion module, as in PWC-Fusion~\cite{ren2019fusion}, or by using a “warm start” initialization where the previous frame’s flow is used to initialize the next estimation, as in RAFT. Recent research has moved beyond pairwise estimation by explicitly modeling sequences of frames. VideoFlow~\cite{shi2023videoflow} introduces a tri-frame optical flow (TROF) module to estimate forward and backward flows from a center frame to its neighboring frames. Multiple TROF modules can then be connected via a motion propagation module to extend to longer video sequences. Another approach, MemFlow~\cite{dong2024memflow}, augments a RAFT-like architecture with a memory buffer that carries forward motion features. StreamFlow~\cite{sun2025streamflow} proposes a streamlined pipeline that processes multiple frames in one forward pass, avoiding redundant calculations of feature maps and correlation volumes. Unfortunately, all three of these approaches do not address the inherent limitations of the cost volume framework's large memory consumption on modern high-resolution videos.

\noindent\textbf{Memory-efficient optical flow.} Memory efficiency has become a critical concern in optical flow estimation since the introduction of RAFT. Methods like Flow1D \cite{xu2021high} and MeFlow \cite{xu2023memory} have explored low-dimensional representations of the correlation volume. Similarly, Sparse Cost Volume (SCV) \cite{jiang2021learning_SCV} restricts the correspondence search of RAFT to a few top matches. On the other hand, Deep Inverse Patchmatch (DIP)~\cite{zheng2022dip} uses a PatchMatch~\cite{barnes2009patchmatch}-based approach to avoid building the all-pairs correlation volume. While these approaches achieve notable improvements in efficiency, they often sacrifice accuracy, falling short of the performance achieved by state-of-the-art methods in the RAFT family. This trade-off between memory efficiency and accuracy highlights the need for novel approaches that can bridge the gap between these competing objectives.

Notably, there has been little work that effectively applies memory-efficient techniques to multi-frame optical flow estimation. In this work, we address this gap by proposing a novel method that enables high-quality optical flow estimation without excessive memory demands.

%% file: 3_method.tex
\begin{figure*}[t]
    \includegraphics[width=\textwidth]{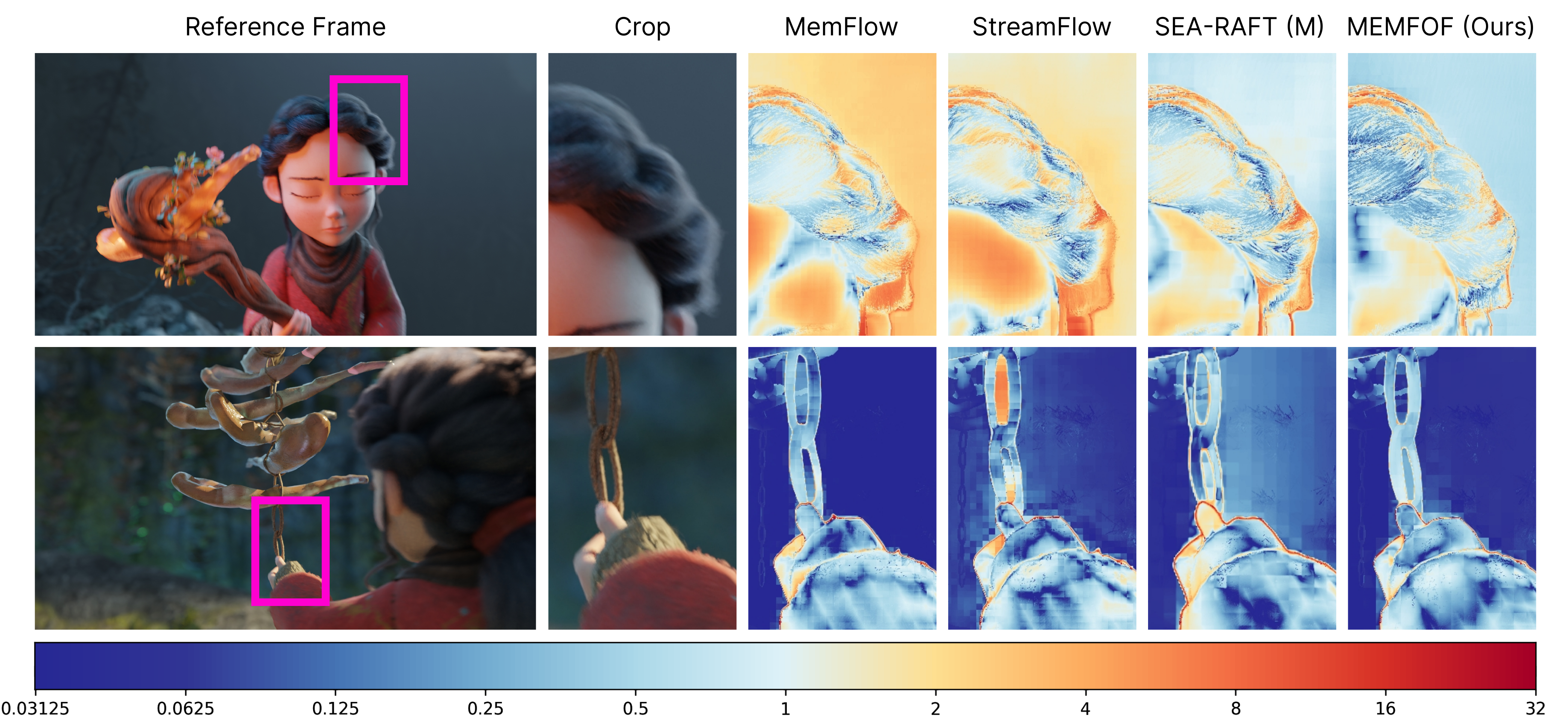}
    \caption{Qualitative comparison of MemFlow~\cite{dong2024memflow}, StreamFlow~\cite{sun2025streamflow}, SEA-RAFT~\cite{wang2025sea}, and our method on Spring benchmark~\cite{mehl2023spring} crops, colorbar represents endpoint error. Our approach surpasses prior methods and demonstrates that: (1) multi-frame processing enhances temporal coherence, and (2) native Full HD resolution preserves local and global motion details. Crops are sourced from official leaderboard submissions.}
    \label{fig:spring-comparison}
\end{figure*}

\section{Method}

Our method introduces a novel approach to optical flow estimation that combines memory efficiency with multi-frame processing without sacrificing accuracy. The method consists of three key components: (1) extending SEA-RAFT to three frames, (2) resolution reduction of the correlation volume, and (3) performance optimization techniques. Below, we describe each component in detail.

\subsection{Extending SEA-RAFT to three frames}
 To leverage temporal information, we extend the two-frame SEA-RAFT architecture to three frames. Following VideoFlow~\cite{shi2023videoflow}, we predict bidirectional flows, one between the current frame and the previous frame, and another between the current frame and the next frame. This involves calculating two correlation volumes instead of one. The update block is also modified to refine both flows at the same time, enabling the network to capture long-range dependencies. Similar to SEA-RAFT, to predict the initial flow, we pass all three frames into the context network. We will now formalize our method.
 
\noindent\textbf{Approach.}
\label{sec:approach}
Given three consecutive frames \(I_{t-1}, I_t, I_{t+1}\), we iteratively estimate a sequence of bidirectional flows \(f^0, f^1, \ldots, f^N  \in (\mathbb{R}^{H \times W \times 2}, \mathbb{R}^{H \times W \times 2}) \); where $N$ indicates the number of iterative refinements; \(f^k\) includes a flow to the previous frame \(f^k_{t \to t-1}\) and a flow to the next frame \(f^k_{t \to t+1}\); $H$ and $W$ are the reduced height and width of the input images.
We begin by extracting the input frame feature maps \(F_t, F_{t-1}, F_{t+1} \in \mathbb{R} ^ {H \times W \times D_f}\). To get the initial prediction \(f^0\), the hidden state \(h^0 \in \mathbb{R}^{H \times W \times D_c}\), and the context features \(g \in \mathbb{R}^{H \times W \times D_c}\), we pass all three frames into the context network:
\begin{align}
g, h^0 &= \text{ContextNetwork}(I_{t-1}, I_t, I_{t+1}), \\
f^0 &= \text{FlowHead}(h^{0}).
\end{align}
The dual correlation volumes \(C_{t,t-1}\) and \(C_{t,t+1}\) are computed as:
\begin{align}
C_{t,t-1}(u, v) &= \langle F_t(u), F_{t-1}(v) \rangle, \\
C_{t,t+1}(u, v) &= \langle F_t(u), F_{t+1}(v) \rangle,
\end{align}
where  \(\langle \cdot, \cdot \rangle\) denotes the dot product.

\begin{table*}[t]
    \caption{Details of our training procedure. Dataset abbreviations: T: Things, S: Sintel, K: KITTI-2015, H: HD1K. Following SEA-RAFT, the dataset distribution for the TSKH stage is S(.32), T(.31), K(.12), H(.24). N indicates the number of iterative refinements used in our method during training. Memory usage is stated per GPU.}
    \label{tab:training}
    \centering
    \input{training}
\end{table*}

\noindent\textbf{Iterative refinement.}
The correlation values \(c^k_{t \to t-1}\) and \(c^k_{t \to t+1}\) are retrieved from the dual correlation volumes based on the current flow predictions:
\begin{align}
c^k_{t \to t-1} &= \text{LookUp}(C_{t,t-1}, f^k_{t \to t-1}), \\
c^k_{t \to t+1} &= \text{LookUp}(C_{t,t+1}, f^k_{t \to t+1}).
\end{align}
These values are then fused and encoded into correlation and flow features, which are in turn transformed into a bidirectional motion feature \(F^k_m\):
\begin{align}
F^k_{\text{corr}} &= \text{CorrEncoder}(c^k_{t \to t-1}, c^k_{t \to t+1}), \\ F^k_{\text{flow}} &= \text{FlowEncoder}(f^k_{t \to t-1}, f^k_{t \to t+1}), \\
F^k_m &= \text{MotionEncoder}(F^k_{\text{corr}}, F^k_{\text{flow}}).
\end{align}
The hidden state \(h^k\) is updated iteratively using the motion feature, context features \(g\), and previous hidden state:
\begin{equation}
h^{k+1} = \text{Updater}(F^k_m, g, h^k),
\end{equation}
and the residual flows \(\Delta f^k\) are decoded from the updated hidden state:
\begin{equation}
\Delta f^k = \text{FlowHead}(h^{k+1}).
\end{equation}
The flow predictions are refined as:
\begin{align}
f^{k+1}_{t \to t-1} &= f^k_{t \to t-1} + \Delta f^k_{t \to t-1}, \\ f^{k+1}_{t \to t+1} &= f^k_{t \to t+1} + \Delta f^k_{t \to t+1}.
\end{align}
The final flow predictions are convexly upsampled to the input resolution as in RAFT.

\subsection{Resolution reduction of the correlation volume}
 A major bottleneck in modern optical flow methods, such as RAFT and SEA-RAFT, is the memory consumption of the correlation volume, which scales quadratically with the input resolution as $\mathcal{O}((HW)^2)$. To address this, we propose reducing the resolutions of the correlation volumes and the working flow predictions to 1/16 of the input frames, compared to the standard 1/8 resolution.

Our three-frame setup benefits from this reduction, decreasing the memory footprint for two correlation volumes from 10.4 GB to just 0.65 GB. While other components (\textit{e.g.}, feature maps and intermediate activations) also contribute to memory usage, preventing a sixteen-fold reduction in overall consumption, the total memory usage remains significantly lower than that of the original two-frame SEA-RAFT (8.19 GB vs. 2.09 GB for FullHD).

To account for the correlation volume size reduction, we adapt the ResNet34~\cite{he2016deep} backbone used in SEA-RAFT. Specifically, to get 1/16 resolution features, we apply a strided convolution on the original 1/8 resolution feature maps. Additionally, to account for more information being stored in each pixel, we increase the feature map dimension $D_f$ from 256 to 1024 and the update block dimension $D_c$ from 128 to 512.

This reduction in memory usage enables training our method in native FullHD, alleviating the need for cropping or downsampling of inputs. Memory consumption during different training stages can be seen in Table~\ref{tab:training}.

\subsection{Performance optimization techniques } 

To further enhance motion coherence, we reintroduce the GMA module~\cite{jiang2021learning_GMA}. To better adapt to different resolutions, similar to MemFlow~\cite{dong2024memflow}, we modify the scale factor in attention from $1 / \sqrt{D_c}$ to $\log_3{(HW)} / \sqrt{D_c}$. 

We additionally apply three inference-time speed and memory optimizations. Firstly, similar to StreamFlow~\cite{sun2025streamflow}, we note that when optical flow needs to be predicted for a video sequence, already calculated feature maps can be reused for future predictions. Secondly, following Flow1D~\cite{xu2021high}, we use convex upsampling only on the last predictions. And finally, we reuse the previously computed correlation volume $C_{t,t+1}$ for overlapping frame pairs when moving to the next frame in video sequence, instead of recomputing it from scratch.

%% file: training.tex
\begin{tabular}{cccccccccc}
\toprule
Stage & Weights & Datasets & Scale & Crop size & N & Learning rate & Batch size & Steps & Memory (GB)\\

\midrule
TartanAir & - & TartanAir & 2x & [480, 960] & 4 & 1.4e-4 & 64 & 225k &  12.0\\
Things & TartanAir & T & 2x & [864, 1920] & 4 & 7e-5 & 32 & 120k & 17.1\\
TSKH & Things & T+S+K+H & 2x & [864, 1920] & 4 & 7e-5 & 32 & 225k & 17.1 \\
Sintel-ft & TSKH & S & 2x & [872, 1920] & 8 & 3e-5 & 32 & 12.5k & 22.8\\
KITTI-ft & TSKH & K & 2x & [750, 1920] & 8 & 3e-5 & 32 & 2.5k & 19.6 \\
Spring-ft & TSKH & Spring & 1x & [1080, 1920] & 8 & 4.8e-5 & 32 & 60k & 28.5 \\

\bottomrule
\end{tabular}

%% file: 4_experiments.tex
\begin{table*}[t!]
    \caption{Benchmark comparison of optical flow methods. Results are sourced from official leaderboard of the Spring benchmark, where minus ("-") indicates the method has no published results. Speed (runtime) and peak GPU memory consumption were measured on a Nvidia RTX 3090 GPU (24 GB) without automatic mixed precision or memory efficient correlation volumes. Lower values are better ($\downarrow$) except for WAUC ($\uparrow$). The best results are indicated in \textbf{bold}, second-best are \underline{underlined}. Method configurations are taken from submissions to the Spring benchmark if present, and from submissions to the Sintel benchmark otherwise.}
    \label{tab:spring-comparison}
    \centering
    \input{spring-comparison}
\end{table*}

\section{Experiments}

We evaluate our method on three popular optical flow benchmarks: Spring~\cite{mehl2023spring} (modern high-resolution sequences), Sintel~\cite{butler2012naturalistic} (synthetic scenes with complex motion) and KITTI~\cite{menze2015object} (autonomous driving).

\subsection{Training Details}

We follow the SEA-RAFT training protocol with some adjustments. We train our method on 32 A100 GPUs with automatic mixed precision. Our main changes with respect to SEA-RAFT are skipping FlyingChairs~\cite{dosovitskiy2015flownet} due to its two-frame limitation, 2$\times$ upsampled frames and flows on datasets other than Spring, and in turn larger crop sizes. Training details are provided in Table~\ref{tab:training}.  In cases when the crop size is bigger than the frame size or is not a multiple of 16, we pad the images with black pixels. Training our main model on all stages takes from 3 to 4 days.
 
\noindent\textbf{Evaluation metrics.} We adopt widely used metrics from established benchmarks~\cite{geiger2013vision, mehl2023spring, richter2017playing} in this study: endpoint error (EPE), 1-pixel outlier rate (1px), Fl-score, and WAUC error. The 1px outlier rate measures the percentage of pixels where the flow error exceeds 1 pixel. The endpoint error (EPE) is defined as the average Euclidean distance between predicted and ground truth flow vectors. The Fl-score measures the percentage of pixels where the disparity or flow exceeds 3 pixels and 5\% of its true value. Finally, the WAUC metric evaluates the inlier rates for a range of thresholds, from 0 to 5 px, and integrates these rates, giving higher weight to lower-threshold rates. Please refer to the supplementary for a formal definition of WAUC.

\noindent\textbf{Mixture-of-Laplace Loss.}
Following SEA-RAFT~\cite{wang2025sea}, we use a mixture-of-Laplace (MoL) loss instead of an L1 loss. The MoL loss for $T$ optical flow frame predictions with $N$ iterative refinements is defined as:
\begin{equation}
    \mathcal{L} = \frac{1}{T} \sum_{t=1}^{T} \sum_{k=0}^{N} \gamma^{N - k} \mathcal{L}^{t,k}_{MoL},
\end{equation}
where \( \mathcal{L}^{t,k}_{MoL} \) is the MoL loss for the $t$-th optical flow frame prediction after $k$ refinement iterations and $\gamma$ is set to 0.85 to add higher weights on later predictions following RAFT. Please refer to the supplementary for more details.

\subsection{Results}

We will now state our results on established public benchmarks.

\noindent\textbf{Results on Spring.} Our approach fine-tunes on and processes native 1080p sequences, which allows it to preserve fine motion details as shown in Figure~\ref{fig:spring-comparison}. This enables state-of-the-art accuracy~--- we outperform SEA-RAFT (M) by 10\% in 1px outlier rate and 2\% in EPE (Table~\ref{tab:spring-comparison}). Additionally, our upsampled pre-train strategy also places us first among all non-fine-tuned submissions, even outperforming the fine-tuned SEA-RAFT (M) by 2.3\% on the 1px metric. Crucially, our memory efficiency allows three-frame temporal processing at native 1080p even with a low memory budget, and our method is faster than other multi-frame competitors.

\begin{table}[t]
    \caption{Evaluation of our method on the Sintel and KITTI-15 public benchmarks. The Sintel benchmark uses EPE as it's metric for both splits, while KITTI-15 uses the Fl-all outliers metric.}
    \label{tab:abl:sintel-kitti}
    \centering
    \input{sintel-kitti}
\end{table}

\begin{table*}[t]
    \caption{Ablation. We validate our training design choices on the Spring training set after the TSKH stage. We compare training at original scales and inference at half scale (baseline) to inference at full resolution and training on either crops of or on full upsampled images. We also study the effect of uni-/bi-directional flow prediction. Our final method is highlighted \colorbox{lightgray}{in gray}. For more details see Sec. \ref{sec:ablation}.}
    \label{tab:abl:train_arch}
    \centering
    \input{train-architecture-ablation}
\end{table*}

\noindent\textbf{Results on Sintel and KITTI.} 
Due to pre-training on 2x upsampled frames, for submissions to the Sintel and KITTI benchmarks, we bilinearly upscale all input images by a factor of two and bilinearly downscale all resulting flow maps by a factor of two. For Sintel submissions we use 16 update iterations. Our method leads on Sintel clean split, surpassing the five-frame version of VideoFlow and outperforms SEA-RAFT (L) by 27\% on the final pass (Table~\ref{tab:abl:sintel-kitti}). On the KITTI benchmark, we achieve state-of-the-art performance among all non-scene flow methods. Please refer to the supplementary material for visual and zero-shot comparisons with other methods.

\subsection{Ablation Study}
\label{sec:ablation}

The ablation study is conducted on the Spring training set (only on the forward left 4K flow), as we mainly focus on FullHD performance.
If not otherwise stated, we use the same training procedure and hyperparameters as in the experiments section --- models after the TSKH stage but before Spring fine-tuning, and perform 8 iterative refinements.

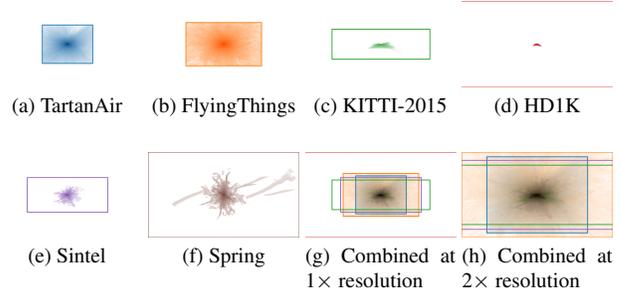
\begin{figure}[t]
    \centering
    \input{abl_2d_hists}
    \caption{We analyze motion patterns in optical flow datasets using 2D histograms. Each histogram uses the same bins, covering all possible motions at FullHD resolution. Color intensity corresponds to the number of motion vectors in each bin. Borders show the maximum motion range in each dataset. (a -- e) Training datasets histograms at their native resolutions. (f) Motion histogram of the Spring training set, note the large motions, not covered by any of the datasets. (g -- h) Combined motion histograms of training datasets without and with 2$\times$ upsampling.}
    \label{exp:abl:mv_hist}
\end{figure}

\noindent\textbf{High-Resolution Training Analysis.}
Commonly used optical flow datasets come in relatively small resolutions, and methods trained on such data often generalize poorly to motion magnitudes seen in high-resolution inputs. This limits the practical use of optical flow methods, causing input downsampling to a resolution that better matches the training stage \cite{lai2022face}.
See Fig.~\ref{exp:abl:mv_hist} for motion vector histogram which illustrates this discrepancy between common datasets and Spring FullHD motion range. Plese refer to the supplementary for details on histogram creation.

We evaluate three strategies to bridge resolution gaps during training, see Table~\ref{tab:abl:train_arch} for detailed metrics:
\begin{itemize}
    \item \textbf{Native Resolution:} Training on original data yields the worst performance (EPE: \textbf{0.430}), as low-res motion magnitudes mismatch FullHD. We additionally test predicting the flow at half the resolution (like in SEA-RAFT \cite{wang2025sea}), which helps improve EPE at large displacements (s40+) but hinders the methods ability to predicts fine motions as shown by all the other metrics.
    \item \textbf{Upsampled (2$\times$) with Crops:} Training on upsampled data cropped to original resolution helps improve the quality but performs worse than full-frame training, likely due to cropped context limiting very large motion learning.
    \item \textbf{Upsampled (2$\times$) Full Frames:} This achieves the best FullHD results (EPE: \textbf{0.341}), as full-frame upsampled training optimally aligns motion distributions with high-res inference.
\end{itemize}

\noindent\textbf{Three-Frame Flow Estimation Strategy.} 
We compare bidirectional (current-to-previous \& current-to-next) and unidirectional (previous-to-current \& current-to-next) flow estimation. As shown in Table~\ref{tab:abl:train_arch}, bidirectional training improves EPE by \textbf{14.75\%} on Spring train data. We posit this stems from simplified motion boundary learning: bidirectional flows share consistent boundaries of the central frame, whereas unidirectional flows face distinct boundaries for each direction, which makes the task of predicting initial flow much harder for the context network.

\begin{table}
    \caption{Ablation. Correlation volume resolution and number of frames. In all models, we set the feature dimension $D_f$ equal to $2D_c$. Please refer to the supplementary material for additional metrics.}
    \label{exp:abl:corr_frames_table}
    \centering
    \input{correlation_res-num_frames-ablation}
\end{table}

\noindent\textbf{Correlation volume resolution, feature dimension, and number of frames.}
We carefully examine the trade-off between the resolution of the correlation volume and the number of frames. We train 2-, 3- and 5- frame models with either 1/16 or 1/24 correlation volume resolution. We also ablate the use of GMA module alongside the hidden state / feature dimensions. The results are reported in Table \ref{exp:abl:corr_frames_table}. We see consistent gains when increasing the feature dimension and a favorable trade-off between 2- and 3-frame models. We also note the performance degradation when moving from 3 to 5 frames which we attribute to the insufficient size of the context network and recursive module, but leave further analysis to future work.

\begin{table}[t]
    \caption{Ablation. Inference time optimizations.}
    \label{exp:abl:infer_table}
    \centering
    \input{inference_time_ablation}
\end{table}

\noindent\textbf{Inference-Time Optimizations.}
Due to the sliding-window processing of video sequences and iterative nature of RAFT-derived methods we note a few calculation redundancies that can be used to further improve runtime.
\begin{itemize}
    \item \textbf{Feature Network Reuse:} When processing sequential data, parts of the computations may be reused. Notably in a three frame scenario we may reuse 2 out of 3 feature network results. We implement this by caching extracted features and only running the feature network on the new frame. This optimization cannot be applied to the context network as it takes all three frames as input.
    \item \textbf{Late Convex Upsampling:} During training all flow predictions are upsampled for loss computation. However, during inference, we need to apply convex upsampling only on the final iteration, eliminating redundant computations.
    \item \textbf{Fast correlation volume:} The official SEA-RAFT implementation naively computes multi-scale correlation volumes between the feature map of the first image and pooled feature maps of the second image. Instead, we compute the correlation volume once and then pool it multiple times.
    \item \textbf{Correlation volume reuse:} Similar to reusing feature maps, correlation volumes can also be reused. By rearranging axes in $C_{t,t+1}$ and then pooling the result multiple times, we can get $C_{t+1,t}$ without performing any matrix multiplications.
\end{itemize}

These optimizations reduce inference time by over \textbf{22\%} when compared to naive implementations of two variants of our method (Table~\ref{exp:abl:infer_table}).

%% file: spring-comparison.tex
\begin{tabular}{l l c c c c c c c}
\toprule
& \multicolumn{1}{l}{\multirow{2}[1]{*}{Method}} & \multicolumn{1}{l}{\multirow{2}[1]{*}{\#Frames}} & \multicolumn{2}{c}{Inference Cost (1080p)} & \multicolumn{4}{c}{Spring (test)} \\
\cmidrule(lr){4-5}\cmidrule(lr){6-9}
 & & & Memory, GB & Runtime, ms & 1px $\downarrow$ & EPE $\downarrow$ & Fl $\downarrow$ & WAUC $\uparrow$ \\
\midrule
\multirow{12}{*}{\rotatebox{90}{NO FINE-TUNE}}
 & Flow1D~\cite{xu2021high} & 2 & 1.34 & 405 & - & - & - & - \\
 & MeFlow~\cite{xu2023memory} & 2 & 1.32 & 1028 & - & - & - & -  \\
 & PWC-Net~\cite{sun2018pwc} & 2 & 1.41 & 76 & 82.265 & 2.288 & 4.889 & 45.670 \\
 & FlowNet2~\cite{ilg2017flownet2} & 2 & 4.16 & 167 & 6.710 & 1.040 & 2.823 & 90.907 \\
 & RAFT~\cite{teed2020raft} & 2 & 7.97 & 557 & 6.790 & 1.476 & 3.198 & 90.920 \\
 & GMA~\cite{jiang2021learning_GMA} & 2 & 13.26 & 1185 & 7.074 & 0.914 & 3.079 & 90.722 \\
 & FlowFormer~\cite{huang2022flowformer} & 2 & OOM & - & 6.510 & 0.723 & 2.384 & 91.679 \\
 & RPKNet~\cite{morimitsu2024recurrent} & 2 & 8.49 & 295 & \underline{4.809} & 0.657 & \underline{1.756} & 92.638 \\
 \cdashlinelr{2-9}
 & VideoFlow-BOF~\cite{shi2023videoflow} & 3 & 17.74 & 1648 & - & - & - & - \\
 & VideoFlow-MOF~\cite{shi2023videoflow} & 5 & OOM & - & - & - & - & - \\
 & MemFlow~\cite{dong2024memflow} & 3 & 8.08 & 885 & 5.759 & 0.627 & 2.114 & 92.253 \\
 & StreamFlow~\cite{sun2025streamflow} & 4 & 18.97 & 1403 & 5.215 & \underline{0.606} & 1.856 & \underline{93.253} \\
 & MEMFOF (Ours) & 3 & 2.09 & 472 & \textbf{3.600} & \textbf{0.432} & \textbf{1.353} & \textbf{94.481} \\
\midrule
\multirow{6}{*}{\rotatebox{90}{FINE-TUNE}}
 & CrocoFlow~\cite{weinzaepfel2023croco} & 2 & 2.01 & 6524 & 4.565 & 0.498 & 1.508 & 93.660 \\
 & SEA-RAFT~(S)~\cite{wang2025sea} & 2 & 8.15 & 205 & 3.904 & 0.377 & 1.389 & 94.182 \\
 & SEA-RAFT~(M)~\cite{wang2025sea} & 2 & 8.19 & 286 & \underline{3.686} & \underline{0.363} & \underline{1.347} & \underline{94.534} \\
 \cdashlinelr{2-9}
 & MemFlow~\cite{dong2024memflow} & 3 & 8.08 & 885 & 4.482 & 0.471 & 1.416 & 93.855 \\
 & StreamFlow~\cite{sun2025streamflow} & 4 & 18.97 & 1403 & 4.152 & 0.467 & 1.424 & 94.404 \\
 & MEMFOF (Ours) & 3 & 2.09 & 472 & \textbf{3.289} & \textbf{0.355} & \textbf{1.238} & \textbf{95.186} \\
\bottomrule
\end{tabular}

%% file: sintel-kitti.tex
\begin{tabular}{l c c c}
\toprule
\multirow{2}[2]{*}{Method} & \multicolumn{2}{c}{Sintel} & \multicolumn{1}{c}{KITTI-15}\\
\cmidrule(lr){2-3} \cmidrule(lr){4-4}
 & Clean $\downarrow$ & Final $\downarrow$ & Fl-all $\downarrow$ \\
\midrule
PWC-Net~\cite{sun2018pwc} & 3.86 & 5.04 & 9.60 \\
FlowNet2~\cite{ilg2017flownet2} & 4.16 & 5.74  & 10.41 \\
Flow1D~\cite{xu2021high} & 2.238 & 3.806 & 6.27 \\
MeFlow~\cite{xu2023memory} & 2.054 & 3.090 & 4.95 \\
RAFT~\cite{teed2020raft}& 1.609 & 2.855 & 5.10 \\
GMA~\cite{jiang2021learning_GMA} & 1.388 & 2.470 & 5.15\\
SEA-RAFT~(M)~\cite{wang2025sea} & 1.442 & 2.865 & 4.64 \\
SEA-RAFT~(L)~\cite{wang2025sea} & 1.309 & 2.601 & 4.30 \\
FlowFormer~\cite{huang2022flowformer} & 1.159 & 2.088 & 4.68 \\
RPKNet~\cite{morimitsu2024recurrent} & 1.315 & 2.657 & 4.64 \\
CrocoFlow~\cite{weinzaepfel2023croco} & 1.092 & 2.436 & 3.64 \\
DDVM~\cite{saxena2023surprising} & 1.754 & 2.475 & \underline{3.26} \\
\cdashlinelr{1-4}
StreamFlow~\cite{sun2025streamflow} & 1.041 & 1.874 & 4.24 \\
MemFlow~\cite{dong2024memflow} & 1.046 & 1.914 & 4.10 \\
MemFlow-T~\cite{dong2024memflow} & 1.081 & 1.840 & 3.88 \\
VideoFlow-BOF~\cite{shi2023videoflow} & 1.005 & \underline{1.713} & 4.44 \\
VideoFlow-MOF~\cite{shi2023videoflow} & \underline{0.991} & \textbf{1.649} & 3.65 \\
MEMFOF (Ours) & \textbf{0.963} & 1.907 & \textbf{2.94} \\
\bottomrule
\end{tabular}

%% file: train-architecture-ablation.tex
\begin{tabular*}{\textwidth}{@{\extracolsep{\fill}}cccccccc cc cc cc cc@{}}
\toprule
\multicolumn{4}{c}{Configuration} & 
\multicolumn{4}{c}{EPE $\downarrow$} & 
\multicolumn{4}{c}{1px $\downarrow$} & 
\multicolumn{2}{c}{Other} \\
\cmidrule(lr){1-4} \cmidrule(lr){5-8} \cmidrule(lr){9-12} \cmidrule(lr){13-14} 
Flow & 
Train & 
\multirow{2}{*}{Crop} & 
Inf. & 
\multirow{2}{*}{avg} & \multirow{2}{*}{s0-10} & \multirow{2}{*}{s10-40} & \multirow{2}{*}{s40+} & 
\multirow{2}{*}{avg} & \multirow{2}{*}{s0-10} & \multirow{2}{*}{s10-40} & \multirow{2}{*}{s40+} & 
\multirow{2}{*}{WAUC $\uparrow$} & \multirow{2}{*}{Fl $\downarrow$}
 \\
Dir. & 
Scale & 
& 
Scale & 
\multicolumn{4}{c}{} & 
\multicolumn{4}{c}{} & 
\multicolumn{2}{c}{}
 \\
\midrule
Bi- & 1x & $\times$ & 1/2 & 0.402 & 0.177 & 1.047 & \underline{6.843} & 4.300 & 2.491 & 16.877 & 35.401 & 93.840 & 1.260 \\
Bi- & 1x & $\times$ & 1 & 0.430 & 0.165 & 0.857 & 8.858 & 3.232 & \underline{1.755} & 11.628 & 33.903 & 94.230 & 0.984 \\
Bi- & 2x & $\checkmark$ & 1 & \underline{0.378} & 0.166 & \textbf{0.811} & 6.960 & \underline{3.195} & 1.815 & \underline{11.231} & 31.933 & 94.192 & \underline{0.873} \\
\rowcolor{lightgray}
Bi- & 2x & $\times$ & 1 & \textbf{0.341} & \textbf{0.133} & \underline{0.818} & \textbf{6.592} & \textbf{3.061} & \textbf{1.739} & \textbf{11.156} & \textbf{29.423} & \textbf{95.604} & \textbf{0.823} \\
\midrule
Uni- & 2x & $\times$ & 1 & 0.400 & \underline{0.137} & 0.869 & 8.732 & 3.281 & 1.888 & 11.633 & \underline{31.563} & \underline{95.157} & 0.917 \\
\bottomrule
\end{tabular*}

%% file: abl_2d_hists.tex
\centering
\begin{minipage}[t]{0.24\linewidth}
    \centering
    \includegraphics[width=\linewidth]{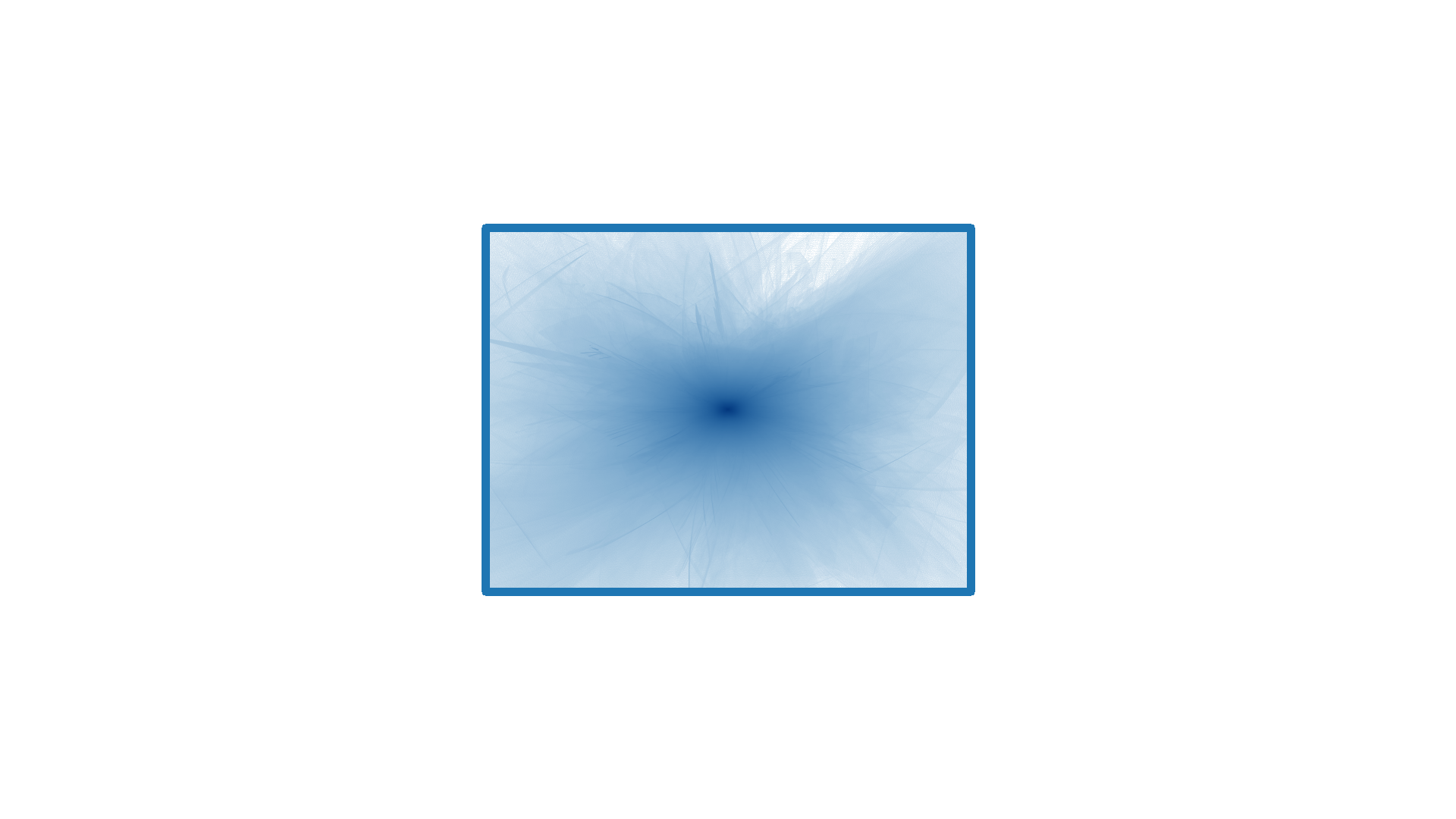}
    \subcaption{TartanAir}
    \label{abl:hists:tartan}
\end{minipage}
\hfill
\begin{minipage}[t]{0.24\linewidth}
    \centering
    \includegraphics[width=\linewidth]{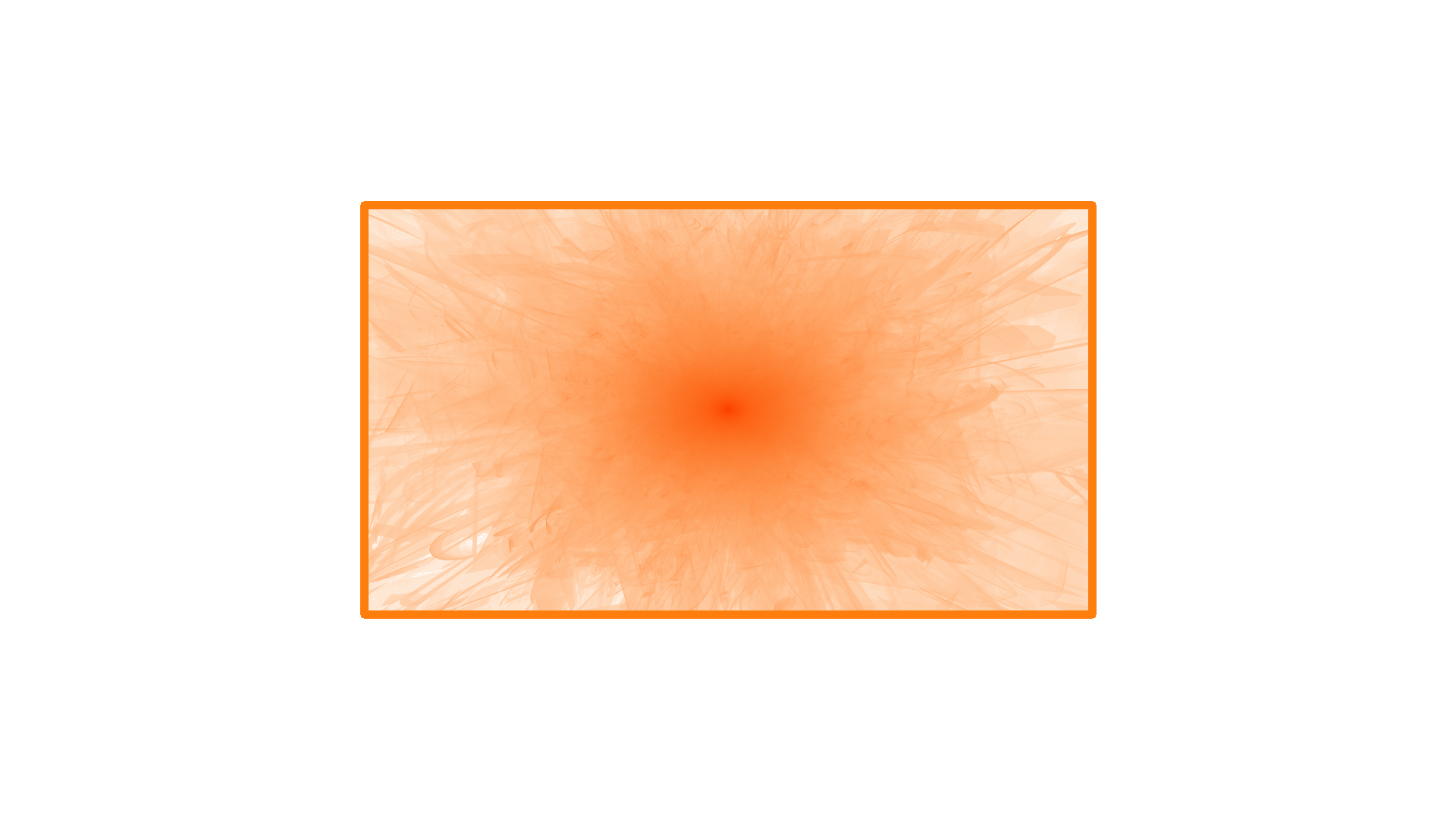}
    \subcaption{FlyingThings}
    \label{abl:hists:things}
\end{minipage}
\hfill
\begin{minipage}[t]{0.24\linewidth}
    \centering
    \includegraphics[width=\linewidth]{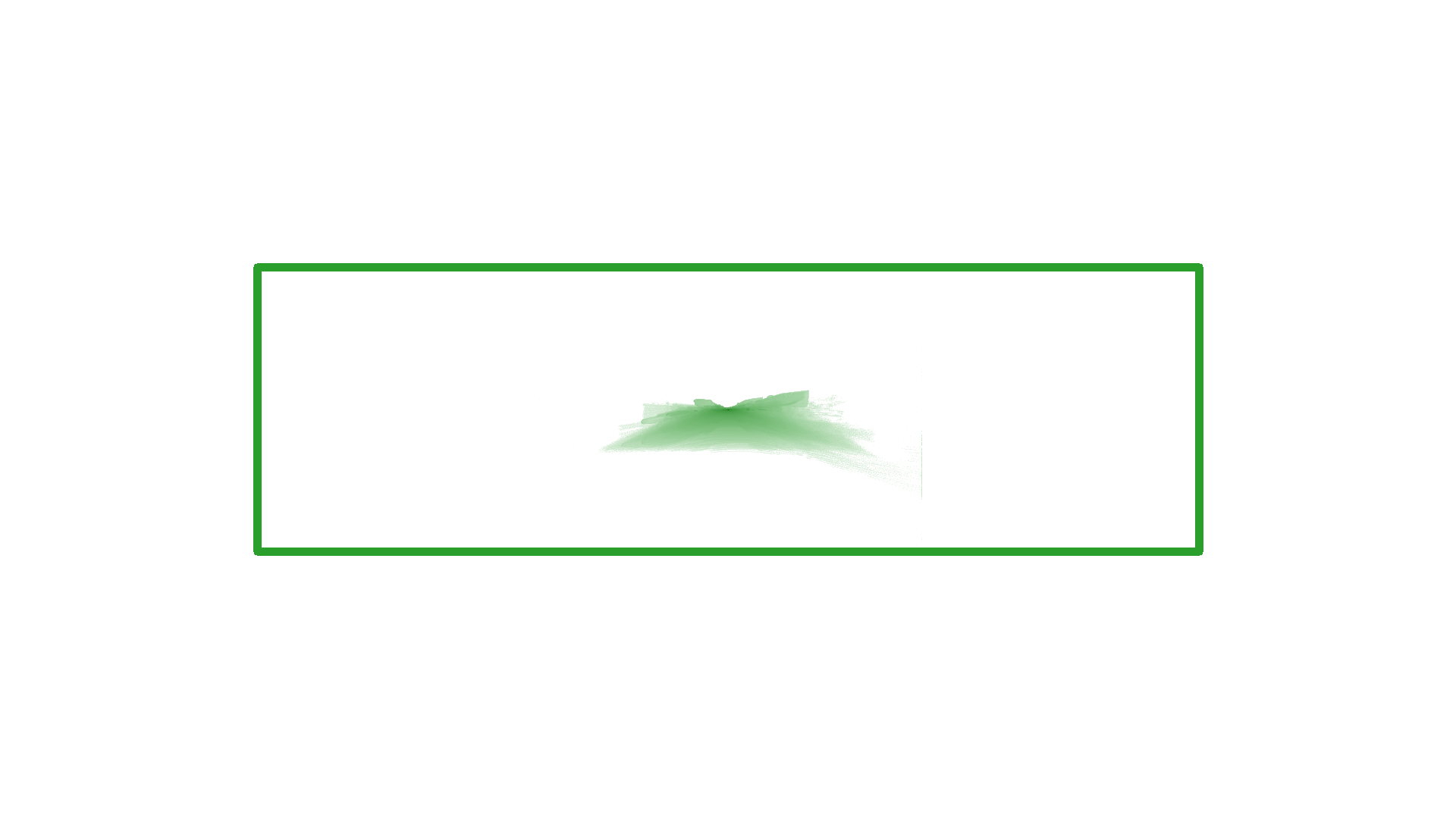}
    \subcaption{KITTI-2015}
    \label{abl:hists:kitti}
\end{minipage}
\hfill
\begin{minipage}[t]{0.24\linewidth}
    \centering
    \includegraphics[width=\linewidth]{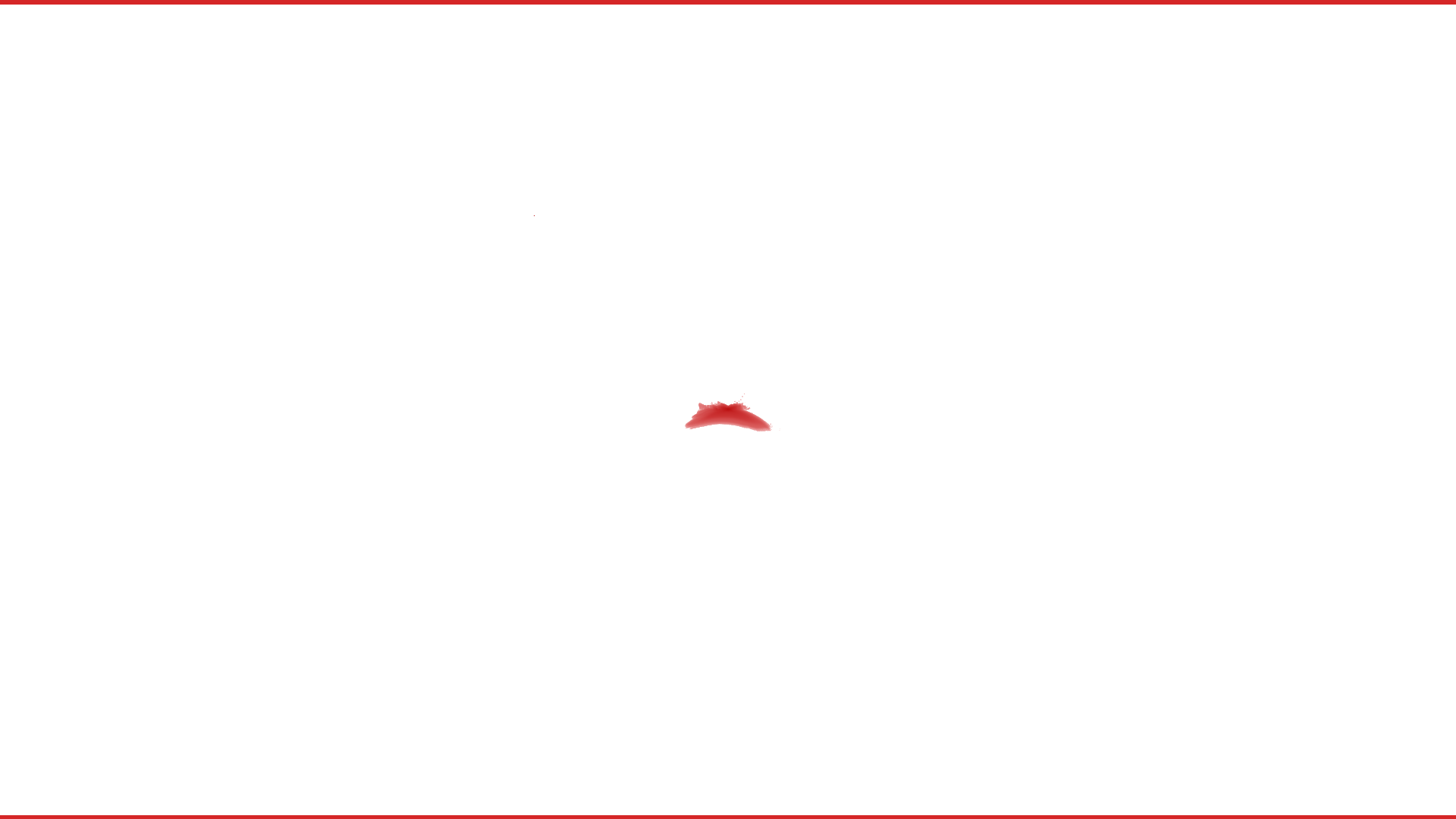}
    \subcaption{HD1K}
    \label{abl:hists:hd1k}
\end{minipage}
\vspace{4mm}

\begin{minipage}[t]{0.24\linewidth}
    \centering
    \includegraphics[width=\linewidth]{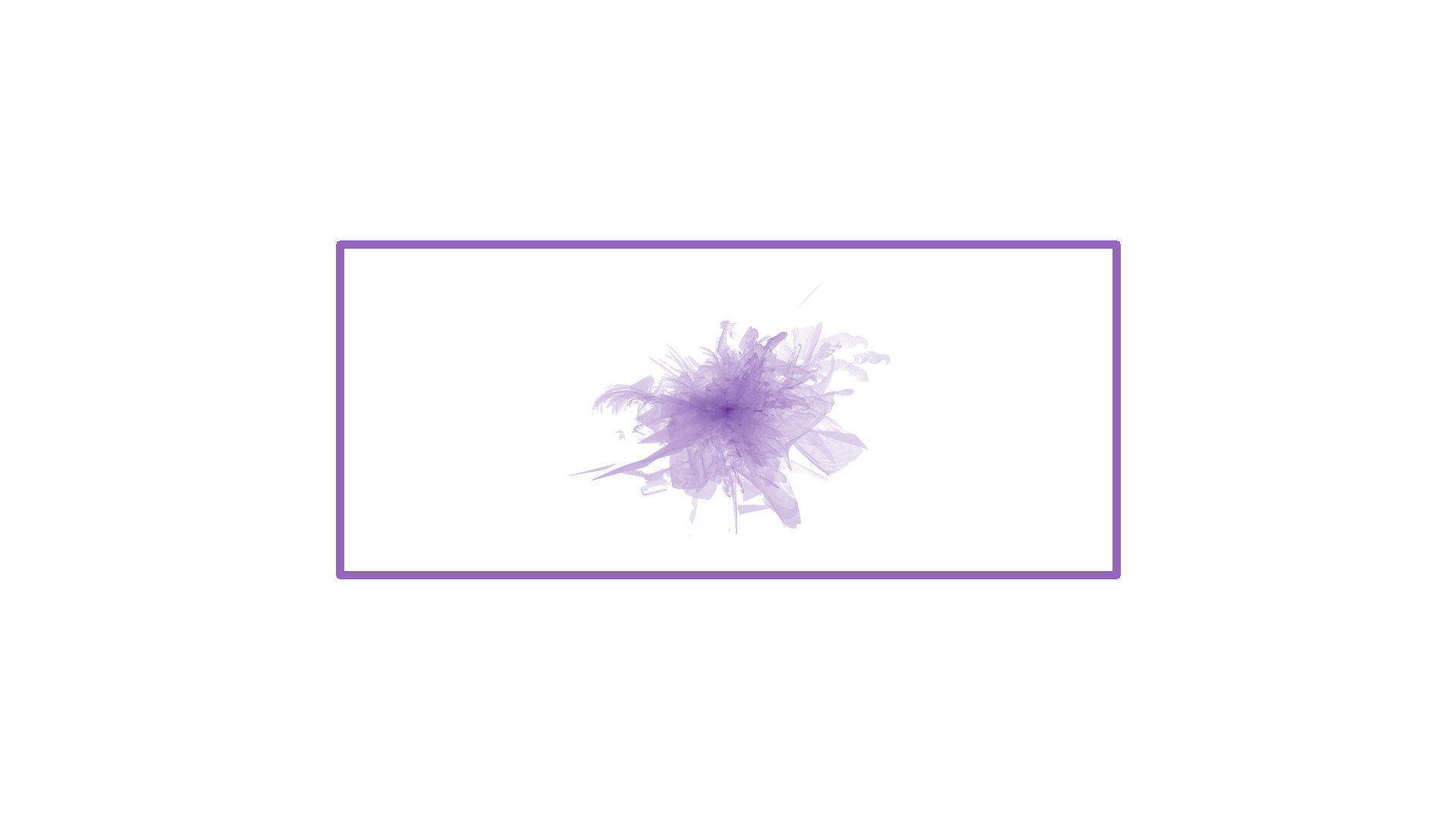}
    \subcaption{Sintel}
    \label{abl:hists:sintel}
\end{minipage}
\hfill
\begin{minipage}[t]{0.24\linewidth}
    \centering
    \includegraphics[width=\linewidth]{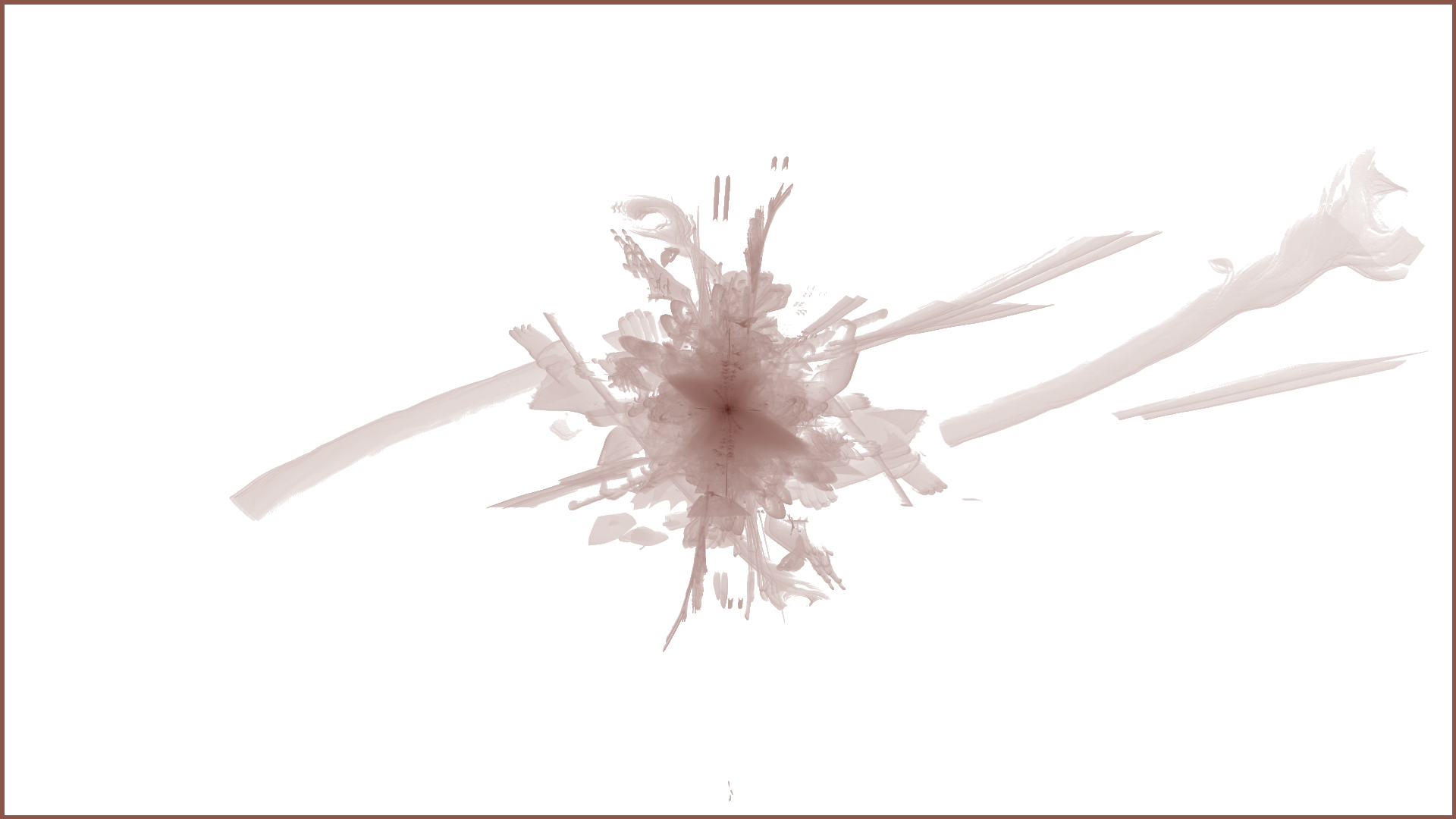}
    \subcaption{Spring}
    \label{abl:hists:spring}
\end{minipage}
\hfill
\begin{minipage}[t]{0.24\linewidth}
    \centering
    \includegraphics[width=\linewidth]{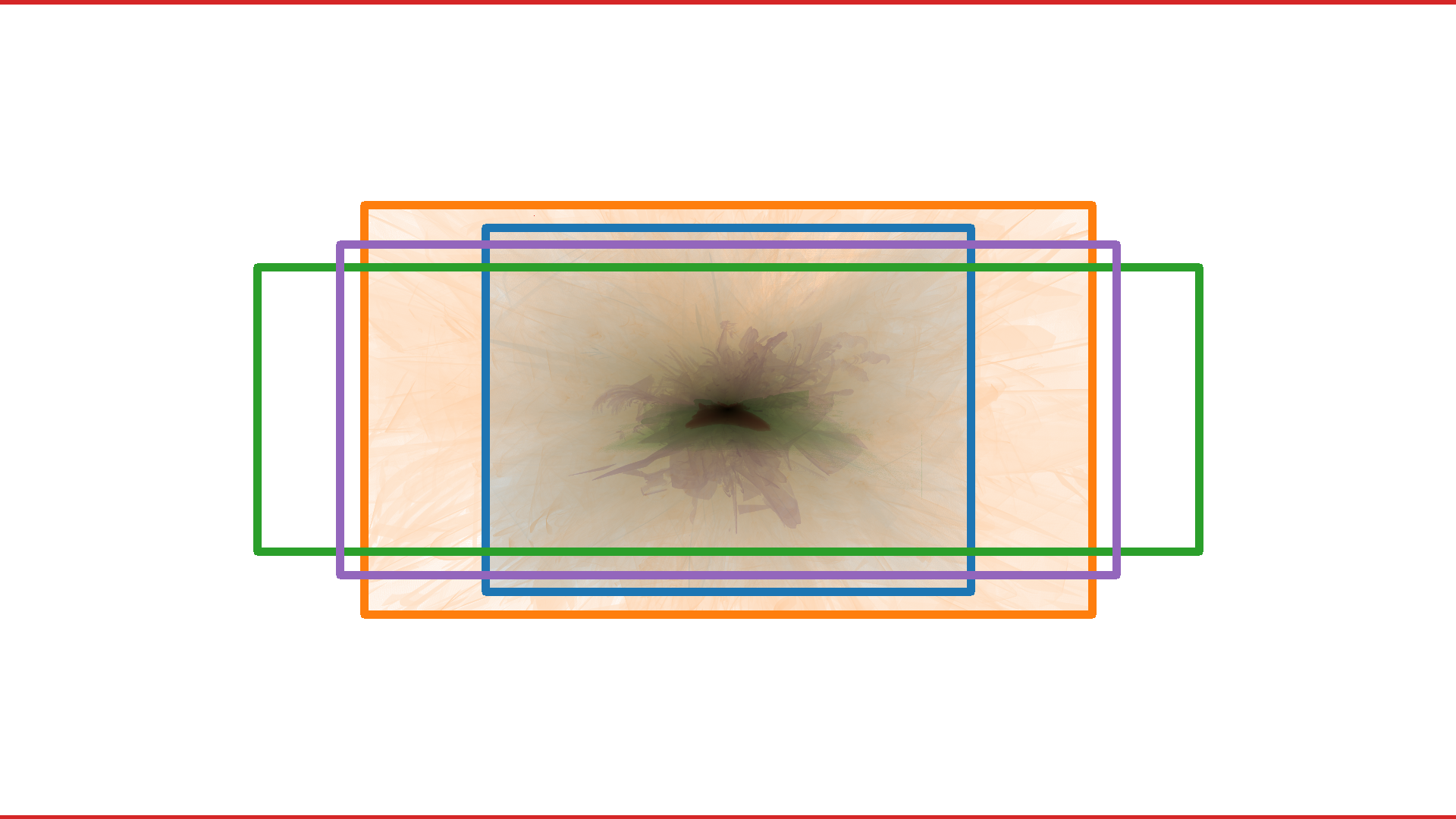}
    \subcaption{Combined at 1$\times$ resolution}
    \label{abl:hists:prod1}
\end{minipage}
\hfill
\begin{minipage}[t]{0.24\linewidth}
    \centering
    \includegraphics[width=\linewidth]{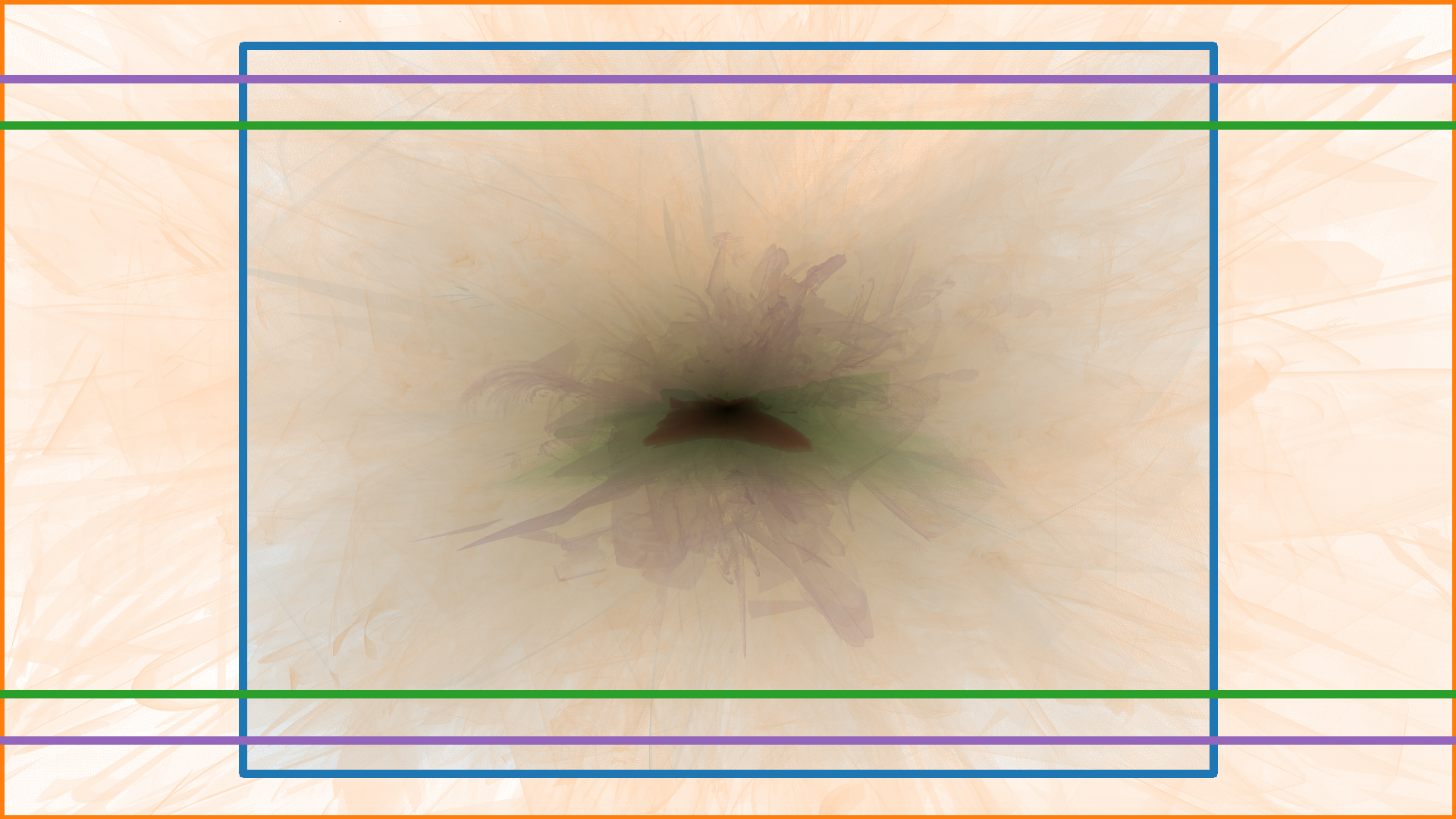}
    \subcaption{Combined at 2$\times$ resolution}
    \label{abl:hists:prod2}
\end{minipage}

%% file: correlation_res-num_frames-ablation.tex
\begin{tabular}{c c c c c c}
\toprule
Corr. scale & \#Frames & $D_c$ & GMA & 1px $\downarrow$ & Mem \\
\midrule
1/24 & 2 & 128 & $\times$ & 4.235 & 0.78 \\
1/16 & 2 & 128 & $\times$ & 3.644 & 1.11 \\
1/16 & 2 & 128 & $\checkmark$ & 3.547 & 1.29 \\
1/16 & 2 & 256 & $\times$ & 3.420 & 1.12 \\
1/16 & 2 & 512 & $\times$ & 3.375 & 1.30 \\
\cdashlinelr{1-6}
1/24 & 3 & 512 & $\checkmark$ & 3.480 & 1.03 \\
1/16 & 3 & 128 & $\checkmark$ & 3.560 & 1.78 \\
1/16 & 3 & 256 & $\checkmark$ & \underline{3.144}& 1.86 \\
\rowcolor{lightgray}
1/16 & 3 & 512 & $\checkmark$ & \textbf{3.061} & 2.09 \\
1/16 & 3 & 512 & $\times$ & 3.151 & 1.82 \\
\cdashlinelr{1-6}
1/24 & 5 & 512 & $\checkmark$ & 3.809 & 1.84 \\
\bottomrule
\end{tabular}

%% file: inference_time_ablation.tex
\begin{tabular}{l c c}
\toprule
\multirow{2}[2]{*}{Method} & \multicolumn{2}{c}{Time, ms } \\
\cmidrule(lr){2-3}
 & 3 fr (1/16) & 5 fr (1/24)\\
\midrule
Baseline & 611 & 597 \\
Only last Convex upsample & 579 & 533 \\
+ Feature network reuse & 483 & 341 \\
+ Fast correlation volume & 478 & 334 \\
+ Correlation volume reuse & 472 & 329 \\
\bottomrule
\end{tabular}

%% file: 5_conclusion.tex
\section{Conclusion}
In this work, we introduced MEMFOF, a memory-efficient multi-frame optical flow method that achieves state-of-the-art performance while maintaining a significantly reduced GPU memory footprint. By systematically revisiting RAFT-like architectures, we identified an optimal trade-off between multi-frame accuracy and memory efficiency, enabling training at native 1080p resolution without the need for cropping or downsampling. Our approach integrates reduced correlation volumes, multi-frame estimation, and high-resolution training strategies to deliver competitive accuracy across multiple benchmarks while operating with lower computational requirements. These findings position MEMFOF as a practical solution for large-scale, high-resolution optical flow estimation, bridging the gap between accuracy and efficiency. Future work may further explore extending our approach to even higher resolutions and real-time applications.

%% file: 6_acknowledgments.tex
\section{Acknowledgments}
This work was supported by the The Ministry of Economic Development of the Russian Federation in accordance with the subsidy agreement (agreement identifer
000000C313925P4H0002; grant No 139-15-2025-012).
The research was carried out using the MSU-270 supercomputer of Lomonosov Moscow State University.
We want to additionally thank Sergey Lavrushkin, Andrey Moskalenko, Ekaterina Shumitskaya and Vladislav Pyatov for proofreading and providing valuable feedback on the manuscript.

%% file: X_suppl.tex
\clearpage
\setcounter{page}{1}
\maketitlesupplementary

\section{Definitions}
\label{sec:definitions}

Here we will provide more detailed definitions used in the main text.

\subsection{WAUC}
\label{sec:definitions:wauc}

In optical flow, weighted area under curve (WAUC), originally from VIPER~\cite{richter2017playing}, is formally defined as the integral
\begin{align}
\frac{2}{5}\int_0^5 f(x) \cdot \frac{5 - x}{5}\, dx,
\end{align}
where $f(x)$ is equal to the percentage of pixels where the flow error does not exceed $x$ pixels. The metric ranges from 0 at worst to 100 at best.

\subsection{Mixture-of-Laplace Loss}
\label{sec:definitions:mol}

For a single flow vector coordinate, the Mixture-of-Laplace (MoL) in SEA-RAFT is defined as:
\begin{multline}
    \text{MixLap}(\mu_{gt}; \alpha, \beta, \mu) = -\log \bigl[ \frac{\alpha}{2} \cdot e^{-|\mu_{gt}-\mu|} + \\ + \frac{1-\alpha}{2e^{\beta}} \cdot e^{-\frac{|\mu_{gt}-\mu|}{e^{\beta}}} \bigr],
\end{multline}
where $\mu_{\text{gt}}$ is the target flow coordinate, $\mu$ is the predicted flow coordinate, $\alpha$ is the predicted mixing coefficient, and $\beta$ is the predicted scale parameter. For a single optical flow frame prediction, the MoL loss is defined as:
\begin{multline}
    \mathcal{L}_{MoL} = \frac{1}{2HW} \sum_{u,v} \sum_{d \in \{x,y\}} \text{MixLap}\bigl(\mu_{\text{gt}}(u,v)_d; \\ \alpha(u,v),
    \beta(u,v), \mu(u,v)_d \bigr).
\end{multline}

\subsection{2D Motion histogram}
In order to visually demonstrate the discrepancy in motion magnitudes between common training datasets and Spring, we construct 2D histograms of motion vectors. Final results can be seen in Figure~\ref{exp:abl:mv_hist}.
The histograms are constructed in the following way:
\begin{equation*}
\begin{split}
    H(u, v) = \sum_{n=1}^{N}\sum_{h=1}^{H}\sum_{w=1}^{W} [u\le f_n(h, w, 0) \le u+1] \\
    \cdot [v\le f_n(h, w, 1) \le v+1],
\end{split}
\end{equation*}
where $f_n\in \mathbb{R}^{H\times W\times2}$ is the nth flow field from a dataset, $(u,v)$ is the motion vector ($u\in[-H', H']$ and $v\in[-W', W']$) and $[\cdot]$ is the Iverson bracket.
We set $H'=1080$, $W'=1920$, therefore our final histograms all have the same $2160\times3840$ resolution, for illustration purposes, we take the logarithm of bin counts. Maximum motion boundaries are derived as twice the size of images in the dataset, since the largest motion possible is to move diagonally from one corner of an image to the other one.

\begin{table}
    \caption{Performance of our main model depending on the number of iterative refinements (N). Metrics are calculated on the Spring train dataset after the TSKH stage. Speed (runtime) was measured on an Nvidia RTX 3090 GPU (24 GB).}
    \label{tab:iters}
    \centering
    \input{iters}
\end{table}

\begin{table}
    \caption{FullHD, method configurations taken from leaderboard sumbissions. Speed (runtime) was measured on an Nvidia RTX 3090 GPU (24 GB).}
    \label{tab:alt_corr}
    \centering
    \input{alt_corr}
\end{table}

\begin{table*}
    \caption{Full correlation volume and number of frames ablation table.}
    \label{tab:abl:corr_full}
    \centering
    \input{correlation_full}
\end{table*}

\begin{table*}[t!]
    \caption{Generalization performance of optical flow estimation on Sintel and KITTI-15 after the "Things" stage. By default, all methods are trained on (FlyingChairs +) FlyingThings3D, additional datasets are listed in the "Extra data" column.}
    \label{tab:abl:c+t}
    \centering
    \input{c+t}
\end{table*}

\section{Additional ablations}

In this section, we provide ablations or ablation data not included in the main text.

\subsection{Number of iterative refinements}

We study our method's behavior depending on the number of iterative refinements. The results are provided in Table~\ref{tab:iters}. For a balance between speed and accuracy, we choose to perform 8 iterative refinements.   

\subsection{Alternative correlation implementation}
We additionally provide memory consumption and speed measurements for RAFT, VideoFlow and our method in Tab. \ref{tab:alt_corr} when using alternative correlation volume implementation that trades compute time for memory efficiency.

\subsection{Corr. volume resolution and number of
frames}

We provide the full version of Table~\ref{exp:abl:corr_frames_table} with additional metrics as Table~\ref{tab:abl:corr_full}.

\section{Additional results}

In this section, we provide some other results that are not included in the main text.

\subsection{Additional zero-shot evaluation}
Following previous works, we evaluate the zero-shot performance of our method after the "Things" training stage on Sintel (train)
and KITTI (train). The results are provided in Table~\ref{tab:abl:c+t}. Our method has the best zero-shot evaluation on KITTI and outperforms SEA-RAFT (L) on Sintel when trained on the same datasets.

\subsection{Qualitative comparison on Sintel and KITTI}

\begin{figure*}
    \includegraphics[width=\textwidth]{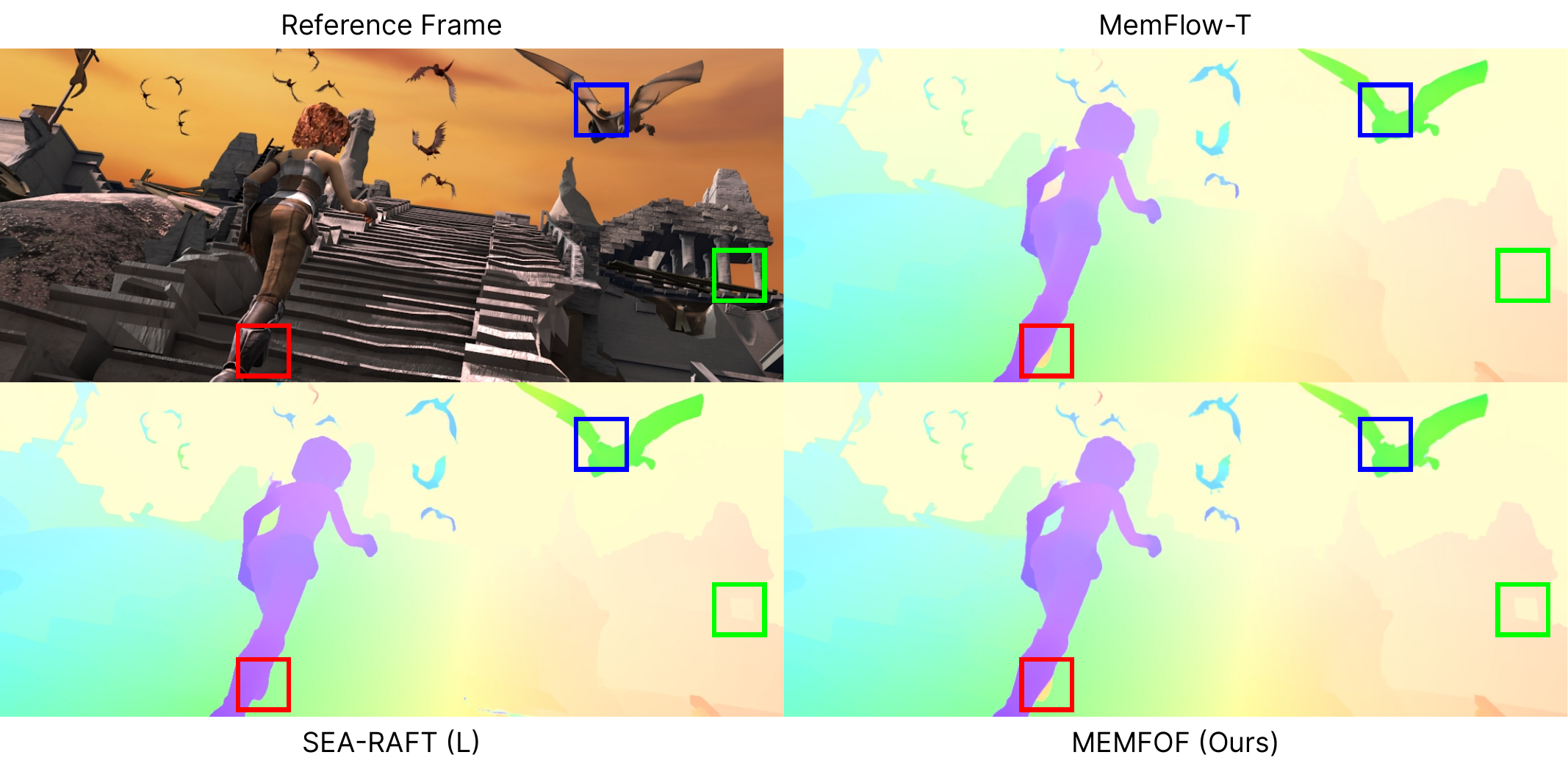}
    \caption{Qualitative comparison of MemFlow-T, SEA-RAFT (L), and our method on the Sintel benchmark. Sourced from official leaderboard submissions.}
    \label{fig:sintel-comparison}
\end{figure*}
\begin{figure*}
    \includegraphics[width=\textwidth]{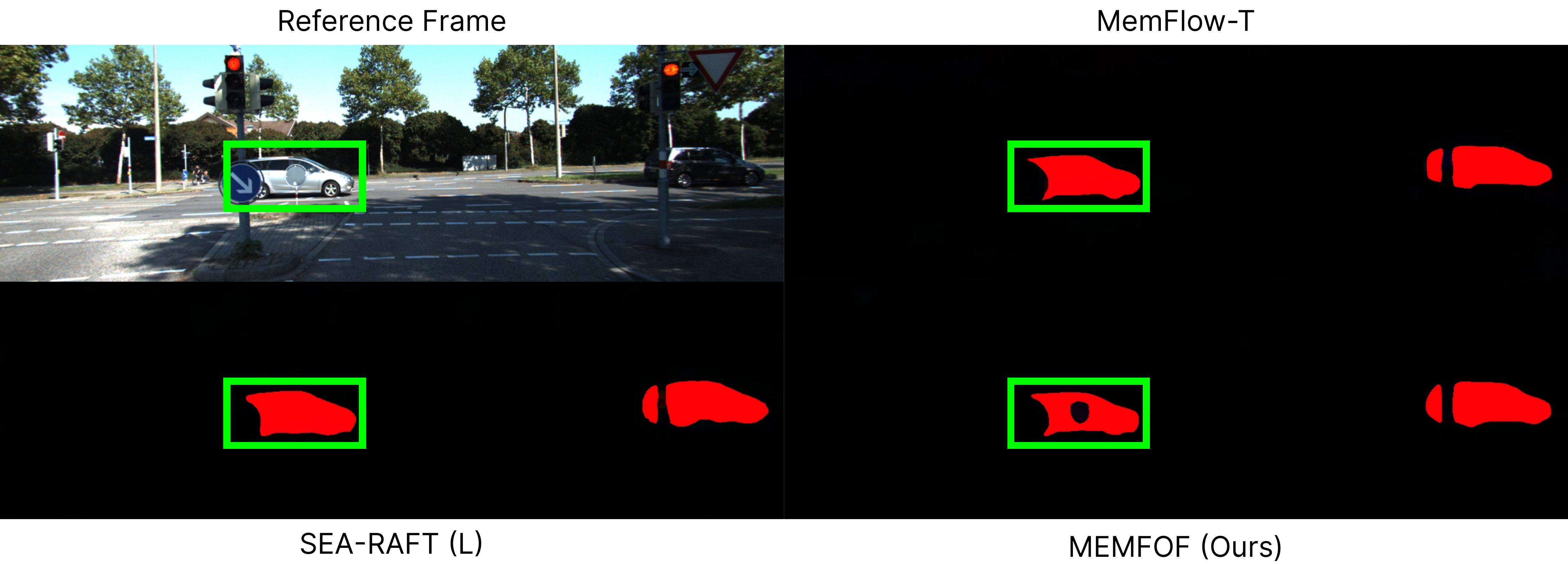}
    \caption{Qualitative comparison of MemFlow-T, SEA-RAFT (L), and our method on the KITTI-2015 benchmark. Sourced from official leaderboard submissions.}
    \label{fig:kitti-comparison}
\end{figure*}

We provide qualitative comparisons of our method on the Sintel and KITTI public benchmarks. As Figure \ref{fig:sintel-comparison} and Figure \ref{fig:kitti-comparison} show, our method has higher motion detail and coherence than our baseline or competitor.

%% file: iters.tex
\begin{tabular}{l c c c c c}
\toprule
N & 1px $\downarrow$ & EPE $\downarrow$ & WAUC $\uparrow$ & Fl $\downarrow$ & Speed, ms\\
\midrule
0 & 6.170 & 0.893 & 90.898 & 2.625 & 71 \\
1 & 3.752 & 0.397 & 94.731 & 1.212 & 172 \\
2 & 3.300 & 0.350 & 95.322 & 0.979 & 215 \\
4 & 3.133 & \textbf{0.339} & 95.565 & 0.863 & 299 \\
6 & 3.081 & \underline{0.340} & \underline{95.603} & 0.835 & 385 \\
\rowcolor{lightgray}
8 & 3.061 & 0.341 & \textbf{95.604} & 0.823 & 472 \\
10 & \underline{3.050} & 0.342 & 95.601 & \underline{0.820} & 557 \\
12 & \textbf{3.045} & 0.342 & 95.598 & \textbf{0.819} & 642 \\
\bottomrule
\end{tabular}

%% file: alt_corr.tex
\begin{tabular}{l c c c c}
\toprule
\multicolumn{1}{l}{\multirow{2}[1]{*}{Method}} & \multicolumn{2}{c}{Standard corr.} & \multicolumn{2}{c}{Alt. corr.} \\
\cmidrule(lr){2-3}\cmidrule(lr){4-5}
 & GB & ms & GB & ms \\
\midrule

RAFT & 7.97 & 557 & 1.32 & 1302 \\
VideoFlow-BOF & 17.74 & 1648 & 7.41 & 3275 \\
MEMFOF & 2.09 & 472 & 1.52 & 1235 \\
\bottomrule
\end{tabular}

%% file: correlation_full.tex
\begin{tabular}{c c c c c c c c c c c c}
\toprule
\multirow{2}[1]{*}{Corr. scale} & \multirow{2}[1]{*}{\#Frames} & \multirow{2}[1]{*}{$D_c$} & \multirow{2}[1]{*}{GMA} & 
\multicolumn{4}{c}{1px $\downarrow$} & \multirow{2}[1]{*}{EPE $\downarrow$} & \multirow{2}[1]{*}{WAUC $\uparrow$} & \multirow{2}[1]{*}{Fl $\downarrow$} & \multirow{2}[1]{*}{Memory, GB} \\
\cmidrule(lr){5-8}
& & & & avg & s0-10 & s10-40 & s40+ & & & & \\
\midrule
1/24 & 2 & 128 & $\times$ & 4.235 & 2.556 & 15.213 & 35.309 & 0.438 & 93.166 &1.150 & 0.78 \\
1/16 & 2 & 128 & $\times$ & 3.644 & 2.232 & 12.171 & 32.141 & 0.396 & 94.574 & 1.167 & 1.11 \\
1/16 & 2 & 128 & $\checkmark$ & 3.547 & 2.132 & 12.101 & 32.025 & 0.408 & 94.617 & 1.035 & 1.29 \\
1/16 & 2 & 256 & $\times$ & 3.420 & 2.072 & 11.440 & 30.941 & 0.372 & 94.761 & 1.018 & 1.12 \\
1/16 & 2 & 512 & $\times$ & 3.375 & 2.047 & 11.201 & 30.614 & 0.350 & 95.130 & 0.888 & 1.30 \\
\cdashlinelr{1-12}

1/24 & 3 & 512 & $\checkmark$ & 3.480 & 1.940 & 13.539 & 32.104 & 0.362 & 94.858 & 0.970 & 1.03 \\
1/16 & 3 & 128 & $\checkmark$ & 3.560 & 2.154 & 12.176 & 31.543 & 0.380 & 94.859 & 1.094 & 1.78 \\
1/16 & 3 & 256 & $\checkmark$ & \underline{3.144} & \underline{1.789} & 11.365 & 30.390 & 0.346 & 95.493 & \underline{0.886} & 1.86 \\
\rowcolor{lightgray}
1/16 & 3 & 512 & $\checkmark$ & \textbf{3.061} & \textbf{1.739} & \underline{11.156} & \textbf{29.423} & \underline{0.341} & \underline{95.604} & \textbf{0.823} & 2.09 \\
1/16 & 3 & 512 & $\times$ & 3.151 & 1.833 & \textbf{10.988} & \underline{30.165} & \textbf{0.332} & \textbf{95.623} & 0.896 & 1.82 \\

\cdashlinelr{1-12}

1/24 & 5 & 512 & $\checkmark$ & 3.809 & 2.164 & 14.389 & 34.620 & 0.408 & 94.546 & 1.117 & 1.84 \\
\bottomrule
\end{tabular}

%% file: c+t.tex
\begin{tabular}{l l c c c c}
\toprule
\multirow{2}[2]{*}{Extra data} & \multirow{2}[2]{*}{Method} & \multicolumn{2}{c}{Sintel (train)} & \multicolumn{2}{c}{KITTI-15 (train)} \\
\cmidrule(lr){3-4}\cmidrule(lr){5-6}
 & & Clean $\downarrow$ & Final $\downarrow$ & Fl-epe $\downarrow$ & Fl-all $\downarrow$ \\
\midrule
 & PWC-Net & 2.55 & 3.93 & 10.4 & 33.7 \\
 & Flow1D & 1.98 & 3.27 & 6.69 & 22.95 \\
 & MeFlow & 1.49 & 2.75 & 5.31 & 16.65 \\
 & RAFT & 1.43 & 2.71 & 5.04 & 17.40 \\
 TartanAir & SEA-RAFT (S) & 1.27 & 3.74 & 4.43 & 15.1 \\
  & SEA-RAFT (M) & 1.21 & 4.04 & 4.29 & 14.2 \\
  & SEA-RAFT (L) & 1.19 & 4.11 & 3.62 & 12.9 \\
\cdashlinelr{1-6}
 & MemFlow & 0.93 & \underline{2.08} & 3.88 & 13.7 \\
& MemFlow-T & \textbf{0.85} & \textbf{2.06} & 3.38 & 12.8 \\
& VideoFlow-BOF & 1.03 & 2.19 & 3.96 & 15.3 \\
& VideoFlow-MOF & 1.18 & 2.56 & 3.89 & 14.2 \\
& StreamFlow & \underline{0.87} & 2.11 & 3.85 & 12.6 \\
& MEMFOF (ours) & 1.10 & 2.70 & \underline{3.31}& \underline{10.08} \\
TartanAir & MEMFOF (ours) & 1.20 & 3.91 & \textbf{2.93} & \textbf{9.93} \\
\bottomrule
\end{tabular}